\documentclass[10pt,twocolumn,letterpaper]{article}

\usepackage{iccv}
\usepackage{times}
\usepackage{epsfig}
\usepackage{graphicx}
\usepackage{amsmath}
\usepackage{amssymb}
\usepackage[accsupp]{axessibility}
\usepackage{algorithm}
\usepackage{algorithmic}
\usepackage{booktabs}
\usepackage{bbding}
\usepackage{pifont}
\usepackage{subfigure}
\usepackage{amsmath}
\usepackage{float}
\usepackage{xcolor}         
\newtheorem{property}{Property}[section]

\usepackage[pagebackref=true,breaklinks=true,letterpaper=true,colorlinks,bookmarks=false]{hyperref}
\usepackage[toc,page]{appendix}

\iccvfinalcopy 

\def\iccvPaperID{8750} 

\ificcvfinal\fi

\begin{document}

\title{Integrating Boxes and Masks: A Multi-Object Framework for\\ Unified Visual Tracking and Segmentation}

\author{
Yuanyou Xu$^{1,2}$\footnotemark[2],\quad Zongxin Yang$^{1}$,\quad Yi Yang$^{1}$\footnotemark[3]\\
$^1$ReLER, CCAI, Zhejiang University, China \space \space \space $^2$Baidu Research, China\\
{\tt\small \{yoxu, zongxinyang, yangyics\}@zju.edu.cn}
}

\maketitle
\ificcvfinal\thispagestyle{empty}\fi

\begin{abstract}

   Tracking any given object(s) spatially and temporally is a common purpose in Visual Object Tracking (VOT) and Video Object Segmentation (VOS). Joint tracking and segmentation have been attempted in some studies but they often lack full compatibility of both box and mask in initialization and prediction, and mainly focus on single-object scenarios. 
   To address these limitations, this paper proposes a Multi-object Mask-box Integrated framework for unified Tracking and Segmentation, dubbed MITS. Firstly, the unified identification module is proposed to support both box and mask reference for initialization, where detailed object information is inferred from boxes or directly retained from masks. 
   Additionally, a novel pinpoint box predictor is proposed for accurate multi-object box prediction, facilitating target-oriented representation learning. 
   All target objects are processed simultaneously from encoding to propagation and decoding, as a unified pipeline for VOT and VOS. 
   Experimental results show MITS achieves state-of-the-art performance on both VOT and VOS benchmarks. Notably, MITS surpasses the best prior VOT competitor by around 6\% on the GOT-10k test set, and significantly improves the performance of box initialization on VOS benchmarks. The code is available at \url{https://github.com/yoxu515/MITS}.
\end{abstract}

\renewcommand{\thefootnote}{\fnsymbol{footnote}}
\footnotetext[2]{Yuanyou Xu worked on this at his Baidu Research internship.}
\footnotetext[3]{Yi Yang is the corresponding author.}
\renewcommand*{\thefootnote}{\arabic{footnote}}

\section{Introduction}
Visual object tracking (VOT) \cite{muller2018trackingnet,fan2019lasot,huang2019got,kristan2022tenth} and video object segmentation (VOS) \cite{pont20172017,xu2018youtube,wang2021survey,li2023transformer} are two critical tasks in computer vision. Visual object tracking involves identifying and tracking specific object(s) in a video stream over time. Video object segmentation aims to segment given object(s) in a video sequence and separate it from the background. Both tasks are essential for applications such as video surveillance and autonomous driving.

\begin{figure}[t]
\includegraphics[width=1.0\columnwidth]{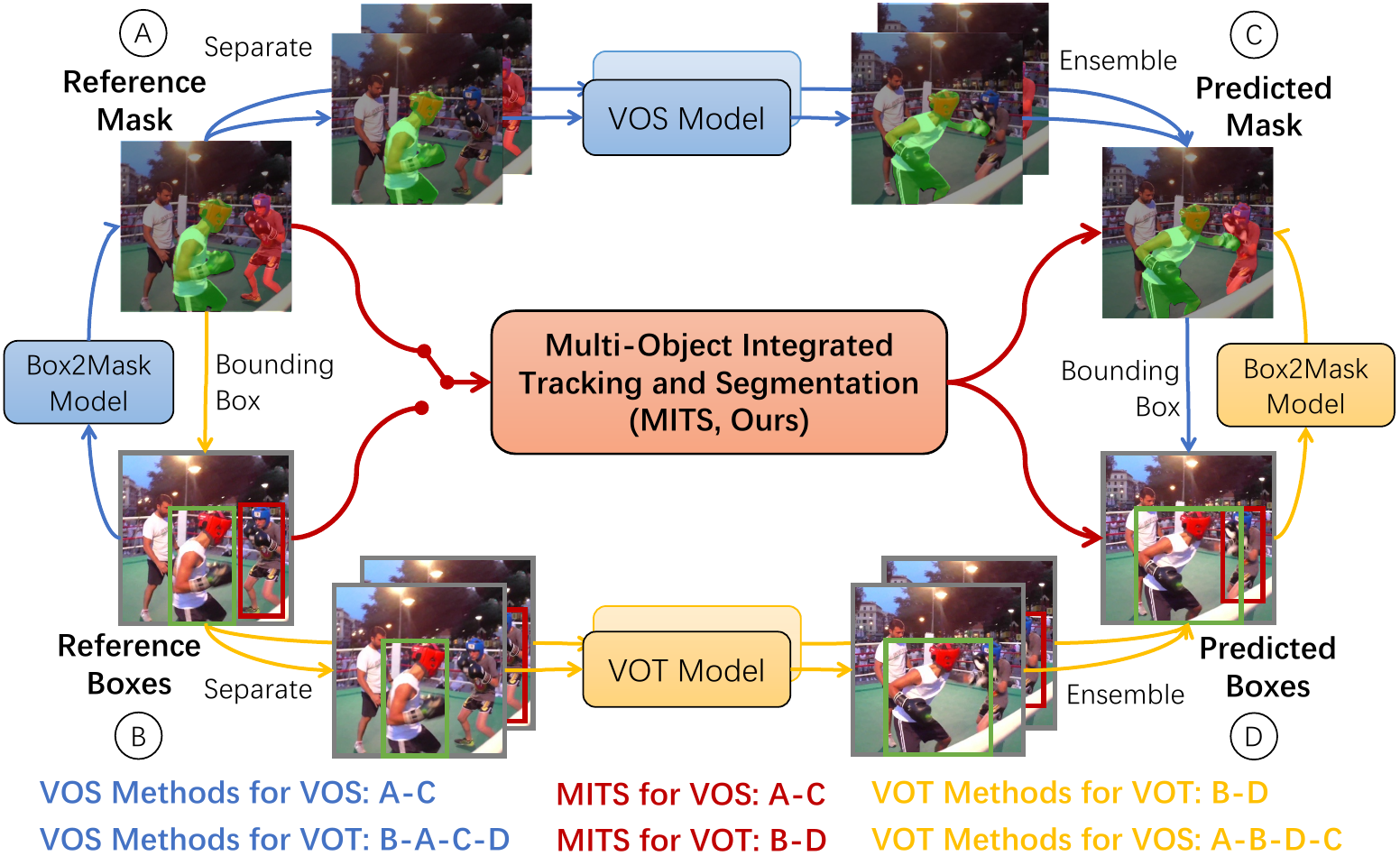}
\caption{We propose a Multi-object Mask-box Integrated framework for unified Tracking and Segmentation (MITS). Compared with VOT or VOS methods, our framework \textit{integrates both boxes and masks} for initialization and prediction in a \textit{unified manner}, without any extra model for box-to-mask conversion, and supports processing \textit{multiple objects} simultaneously for VOT and VOS.}
\label{fig:open}
\end{figure}

In VOT, object sizes and positions are indicated with boxes as box representation, while in VOS object shapes and contours are marked with pixel-level masks as mask representation. Despite their differences, VOT and VOS share similarities. Both tasks require the ability to identify and locate the target object in a video stream accurately in spatial dimension, and to be robust against challenges such as occlusion and fast motion in temporal dimension.

 In view of the similarity, tracking and segmentation may have unified multi-knowledge representations \cite{mkr} and have been explored jointly in some works. 1) \textit{Unification}. A straightforward solution is to perform conversion between boxes and masks to utilize VOT methods on VOS or VOS methods on VOT. A mask can be converted to a box easily, but generating a mask from a box is hard. Some methods were proposed to address the box-to-mask estimation problem \cite{luiten2019premvos,yan2021alpha,zhao2021generating}. However, separate but not unified models hinder end-to-end training and are inconvenient to manage in practical applications. 
 2) \textit{Compatibility}. Several studies have attempted to unify these two tasks into a single framework. However, some of them \cite{wang2019fast,voigtlaender2020siam,paul2022robust} still lack compatibility and flexibility in box/mask input/output and resort to extra models. 
 3) \textit{Multi-Object}. Despite that some methods \cite{lukezic2020d3s,wang2021different,yan2022towards} possess strong compatibility across VOT and VOS, they mainly focus on the single object scenario and use an ensemble strategy to aggregate the separate result of each object in the multiple object scenario.

Therefore, this paper aims to unify VOS and VOT and improve above shortcomings by integrating boxes and masks in a multi-object framework as Multi-object Integrated Tracking and Segmentation (MITS), as shown in Figure \ref{fig:open}. For compatibility problem, a unified identification module is proposed to take both reference boxes in VOT and masks in VOS for initialization. The unified identification module encodes the reference boxes or masks into the unified identification embedding by assigning identities to objects. The coarse identification embedding from boxes is further refined to mitigate the gap between mask and box initialization. The unified identification module is more convenient than borrowing an extra model because it is trained with the whole model in an end-to-end manner.

Besides, the novel pinpoint box predictor is proposed for joint training and prediction with the mask decoder. Previous corner head or center head estimates a box by corners or a center point, which are not have to be inside the object. Emphasizing exterior points may distract learning target-oriented features and affect the mask prediction. To address this problem, we estimate the box by localizing pinpoints, which are always on the edge of the object. However, directly supervise the learning of pinpoints is infeasible due to the lack of annotation. Therefore we perform decoupled aggregation on the pinpoint maps and determine the box only by side-aligned pinpoint coordinates.

All the modules in our framework are not only compatible with two tasks, but also able to process multiple objects simultaneously. The multi-object training and prediction make our framework efficient and robust under complex scenes with multiple objects. Extensive experiments are conducted to demonstrate the strong compatibility and capacity of our framework. Experimental results show that our framework achieves SOTA performance on VOT benchmarks including LaSOT \cite{fan2019lasot}, TrackingNet \cite{muller2018trackingnet} and GOT-10k \cite{huang2019got}, and VOS benchmark YouTube-VOS \cite{xu2018youtube}. Our method improves 6\% over previous SOTA VOT method on GOT-10k, and significantly improves the performance of box initialization on VOS benchmarks.
In summary, our contributions are:
\vspace{-5pt}
\begin{itemize}
    \item We present a multi-object framework integrating boxes and masks for unified tracking and segmentation. \vspace{-3pt}
    \item The unified identification module is proposed to accept both masks and boxes for initialization.\vspace{-3pt}
    \item A novel pinpoint box predictor is proposed for accurate box prediction together with the mask decoder.
\end{itemize}
\section{Related Work}
\paragraph{Visual Object Tracking.} In our context, we use visual object tracking (VOT) as a union of single object tracking (SOT) and multi-object VOT, preventing confusion with multi-object tracking (MOT) \cite{milan2016mot16} which mainly considers object association between two detected object sets. VOT has been well studied in recent years. Correlation filter based methods \cite{bolme2010visual,henriques2014high,danelljan2016beyond,danelljan2017eco,lukezic2017discriminative,bhat2019learning,danelljan2020probabilistic,mayer2022transforming} train a correlation filter on training features and perform convolution on test features to predict classification scores for the target object.
Siamese approaches \cite{bertinetto2016fully,valmadre2017end,li2018high,li2019siamrpn++,guo2020siamcar,voigtlaender2020siam} use a Siamese network to learn a offline similarity metric between the template and search region, and perform cross-correlation to localize the target object.
Recently some works \cite{yu2020deformable,du2020correlation,chen2021transformer,wang2021transformer,yan2021learning,song2022transformer} adopt transformers \cite{vaswani2017attention} in visual tracking for feature extraction and correlation modeling \cite{xie2022correlation,cui2022mixformer,ye2022joint,chen2022backbone}, which achieve promising performance. 
However, most VOT studies mainly focus on SOT, while ours is the first multi-object VOT framework.

\vspace{-5pt}
\paragraph{Video Object Segmentation.} In video segmentation tasks~\cite{li2022video, liang2023local, li2023tube, samtrack}, semi-supervised video object segmentation aims to track given objects with masks rather than boxes in VOT. 
Online VOS methods \cite{caelles2017one,voigtlaender2017online,maninis2018video,yang2018efficient,meinhardt2020make} fine-tune a segmentation model on the reference mask for each video, while matching-based VOS methods \cite{shin2017pixel,oh2018fast,wang2019ranet,voigtlaender2019feelvos,yang2020collaborative,yang2021collaborative} measure the pixel-level similarity to segment target objects. 
Space-time memory network (STM) \cite{oh2019video} leverages a memory network to read object information from past frames with predicted masks, and is further improved and extended by many following works \cite{lu2020video,hu2021learning,seong2021hierarchical,cheng2021rethinking,yang2021associating,zhang2023boosting}. Since masks are more consuming to annotate than boxes, VOS benchmarks \cite{xu2018youtube,pont20172017} usually have shorter videos than SOT benchmarks \cite{fan2019lasot,muller2018trackingnet}. Some methods \cite{li2020fast,liang2020video,cheng2022xmem} have been proposed for long-term VOS, but they lack consideration for VOT. 
AOT \cite{yang2021associating,yang2021towards} and its following works \cite{yang2022decoupling,xu2023video,samtrack} realize simultaneous processing of multiple objects in VOS by the multi-object identification mechanism. We adopt the mechanism in our framework and further improve it for unified tracking and segmentation.

\vspace{-5pt}
\paragraph{Multi-Object Tracking and Segmentation.} MOT \cite{milan2016mot16}/MOTS \cite{voigtlaender2019mots} is quite different from VOT/VOS, and the former relies on object detectors trained on \textit{pre-defined} object categories, and focuses on the association of detected results. Although some work \cite{wang2021different,yan2022towards} unified MOT/MOTS and VOT/VOS, we recognize the large gap between these two types of tasks and consider VOT and VOS that can generalize to \textit{arbitrary} objects as the scope of our work.

\begin{table}[]
\centering
\renewcommand{\arraystretch}{0.8}
\resizebox{\columnwidth}{!}{%
\begin{tabular}{@{}l|cc|cc|c|c@{}}
\toprule
                                      & \multicolumn{2}{c|}{Initialization} & \multicolumn{2}{c|}{Prediction} & \multicolumn{1}{l|}{}            & \multicolumn{1}{l}{} \\
Method                                & Box              & Mask             & Box            & Mask           & Extra Model                      & Multi-Object         \\ \midrule
SiamMask \cite{wang2019fast}          & \ding{51}        & \ding{55}        & \ding{51}      & \ding{51}      & -                                & \ding{55}            \\
D3S \cite{lukezic2020d3s}             & \ding{51}        & \ding{51}        & \ding{55}      & \ding{51}      & -                                & \ding{55}            \\
SiamR-CNN \cite{voigtlaender2020siam} & \ding{51}        & \ding{55}        & \ding{51}      & \ding{55}      & Box2Seg \cite{luiten2019premvos} & \ding{55}            \\
UniTrack \cite{wang2021different}     & \ding{51}        & \ding{51}        & \ding{51}      & \ding{51}      & -                                & \ding{55}            \\
Unicorn \cite{yan2022towards}         & \ding{51}        & \ding{51}        & \ding{51}      & \ding{51}      & -                                & \ding{55}            \\
RTS \cite{paul2022robust}             & \ding{55}        & \ding{51}        & \ding{55}      & \ding{51}      & STA \cite{zhao2021generating}    & \ding{55}            \\ \midrule
MITS (Ours)                           & \ding{51}        & \ding{51}        & \ding{51}      & \ding{51}      & -                                & \ding{51}            \\ \bottomrule
\end{tabular}%
}
\caption{We compare different prior unified approaches for unified visual object tracking and segmentation. For each method, we consider whether it offers native support for box and mask initialization and prediction without any extra model, and whether it is a multi-object framework.}
\label{tab:methods}
\end{table}

\vspace{-5pt}
\paragraph{Unified Tracking and Segmentation.} Prior unified tracking and segmentation methods are listed in Table \ref{tab:methods}. Wang \textit{et al.} developed a unifying approach SiamMask \cite{wang2019fast} by augmenting previous tracking methods \cite{bertinetto2016fully,li2018high} with a segmentation loss. 
D3S \cite{lukezic2020d3s}, a segmentation-centric tracker, applies two target models with geometric properties. The approximate mask is first estimated by a self tracking iteration for box initialization.
Siam R-CNN \cite{voigtlaender2020siam} performs tracking by re-detection \cite{girshick2015fast}. An extra off-the-shelf Box2Seg \cite{luiten2019premvos} network is needed to generate masks from boxes for the segmentation task. Alpha-Refine \cite{yan2021alpha} is a plug-and-play model that refines coarse predictions from a base tracker. It provides good practice for refining tracking results by mining detailed spatial information.

UniTrack \cite{wang2021different} uses a shared backbone for five different tracking related tasks (SOT, VOS, MOT, MOTS, PoseTrack \cite{andriluka2018posetrack}), while its main topic is the self-supervised pre-trained representation for these tasks.
Unicorn \cite{yan2022towards} unifies four tracking tasks by using the target prior in the unified input head, and sets up two branches for box and mask output. UNINEXT \cite{yan2023universal} further unifies more instance perception tasks as object discovery and retrieval. Differently, we only consider VOT and VOS with strong generalization ability for compact unification. RTS \cite{paul2022robust} performs robust tracking by augmenting a VOS method \cite{bhat2020learning} with an instance localization branch. An extra STA \cite{zhao2021generating} network is borrowed to convert a box annotation to a mask for initialization in VOT. Compared with it, we follow the mask-oriented modeling but our framework supports multi-object, more compatible in input and output, and also achieves higher performance.

\section{Method}
\subsection{Preliminary}
\paragraph{Multi-Object Identification Mechanism.}
The multi-object identification mechanism is proposed in the VOS method AOT \cite{yang2021associating} and used in its following work \cite{yang2022associating,yang2022decoupling}. Information about all the target objects in the reference masks is encoded into an IDentification Embedding by an IDentity Bank. In detail, the ID bank stores $M$ learnable high dimensional ID vectors, and each ID is responsible for representing one object. The ID embedding is obtained by assigning ID vectors to $N$ objects ($N\leq M$) according to the reference mask. We adopt this mechanism in our framework, and further improve it by integrating both box and mask representation in a Unified IDentification Module, which achieves unified initialization for VOS and VOT.

\vspace{-5pt}
\paragraph{Pinpoint Definition \& Properties.} \textit{Pinpoints} are defined as the contiguous points between an object and its axis-aligned rectangle bounding box, which pin the box tightly around the object. Pinpoints are first proposed as \textit{Extreme Points} for efficient object annotation instead of boxes \cite{papadopoulos2017extreme}. We use the name \textit{Pinpoints} in our paper to emphasize the connection constructed by pinpoints between the box and mask of an object, and further explore its properties for box head design.
\vspace{-5pt}
\begin{property}[Pinpoint Existence]
\label{prop: p1}
Given an object and its axis-aligned rectangle bounding box, there is at least one pinpoint on each side of the box. (An example for multiple pinpoints on one box side is in Figure \ref{fig: pin_viz}.)
\end{property}

\vspace{-15pt}
\begin{property}[Pinpoint-Based Box]
\label{prop: p2}
An axis-aligned rectangle bounding box can be fully determined by four parameters, which are the side-aligned coordinates of four pinpoints on four sides respectively, i.e. the y coordinates of top and bottom side pinpoints and the x coordinates of left and right side pinpoints (Illustrated in Figure \ref{fig: box head} and \ref{fig: pin_viz}).
\end{property}

\subsection{Pipeline Overview}
\begin{figure*}[t]
\centering
\includegraphics[width=0.95\textwidth]{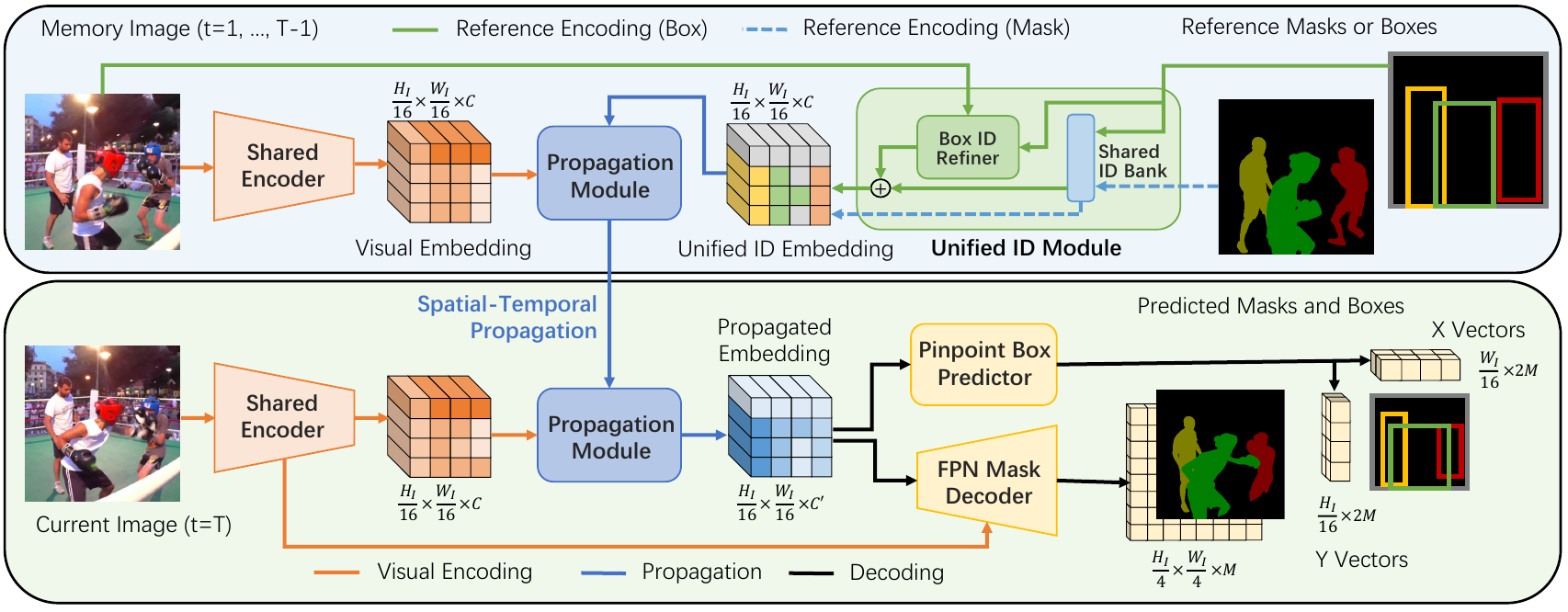}
\caption{The unified encoding-propagation-decoding pipeline in MITS for VOT and VOS task. Both box and mask reference can be encoded by the unified identification module. The unified ID embedding is propagated by propagation module together with the visual embedding from past frames to the current frame. The propagated embedding is decoded into boxes and masks by separate branches.}
\label{fig:MITS}
\end{figure*}
Our unified pipeline for VOT and VOS tasks can be formulated as encoding-propagation-decoding (Figure \ref{fig:MITS}). There are two beliefs for constructing our unified framework. The first is \textit{mask-oriented}, i.e. extracting and maintaining detailed spatial information as much as possible, which has been proved to be more robust in prior work \cite{lukezic2020d3s,yan2021alpha,paul2022robust}. The second is \textit{multi-object mask-box integrated}, i.e. natively supporting simultaneously processing multiple boxes or masks from input to output, which has never been achieved by prior work. 


\vspace{-5pt}
\paragraph{Unified Box-Mask Encoding.} At the beginning, a reference frame or memory frame $I_{m}\in R^{H_I\times W_I}$ is encoded by a backbone encoder (which shares weights for all frames) into the visual embedding $X_{m}\in R^{HW\times C}$. The reference masks $Y_m\in R^{HW\times N}$ or boxes for $N$ objects are encoded by the Unified Identification Module (Section \ref{sec: UIDM}) into the Unified IDentification Embedding $E_{id}\in R^{HW\times C}$. Then, the visual embedding is sent to the propagation module to obtain the memory embedding for propagation, together with the ID embedding. Note that any frame with predicted boxes or masks can become a memory frame, and the memory embedding is extended from $R^{HW\times C}$ to $R^{THW\times C}$ with $T$ memory frames.

\vspace{-5pt}
\paragraph{Spatial-Temporal Propagation.} For the current frame, the visual embedding obtained from the encoder is fed into the propagation module. In the propagation module, the embedding of current frame is transformed into the query $Q_t\in R^{HW\times C}$ and the memory embedding is transformed into the key $K_m\in R^{THW\times C}$ and value $V_m\in R^{THW\times C}$. The ID embedding $E_{id}\in R^{THW\times C}$ is fused with $V_m$ and propagated from memory frames to the current frame via attention mechanism \cite{vaswani2017attention,oh2019video,yang2021associating, yang2022decoupling}:
\begin{align}
Attn&(Q,K,V) = Softmax(\frac{QK^T}{\sqrt{d}}) \cdot V,\\
&V_t = Attn(Q_t,K_m,V_m+E_{id}).\label{eq:prop}
\end{align}

\vspace{-5pt}
\paragraph{Dual-Branch Decoding.} Finally, the embedding after propagation is passed to two decoding branches for different prediction. The decoder for predicting masks is a feature pyramid network (FPN) \cite{lin2017feature}. While the decoder for predicting boxes is a transformer-based pinpoint box head (Section \ref{sec: box head}). Both of these two decoders predict masks or boxes for all objects together. For the mask branch, the predicted logits $S_m\in R^{H_sW_s\times M}$ are for object capacity $M$, and the probability $P_m\in R^{HW\times N}$ for actual $N$ objects is obtained by using $Softmax$ on $N$ channels of $M$. For the box branch, the predicted probability vectors $P_b^{x_1,x_2}\in R^{W\times 2M}$ and $P_b^{y_1,y_2}\in R^{H\times 2M}$ form $M$ boxes. We choose $N$ boxes from $M$ predicted boxes.

\subsection{Unified Identification Module}
\label{sec: UIDM}

For mask-based VOS models, directly taking a box for initialization will lead to severe performance degradation \cite{voigtlaender2020siam,paul2022robust}, as shown in Table \ref{tab:SOT}. A choice is to borrow an extra model which generates masks from boxes \cite{paul2022robust}. We argue such a strategy has two shortcomings. First, it is cumbersome train another extra model. Second, the extra model can't be optimized together with the VOS model. To address these problems, we propose a Unified IDentification Module (UIDM), which has three advantages: 1) It supports both box and mask initialization, keeping rich details in the masks while refining coarse information in the boxes. 2) It encodes the reference in a multi-object way, all objects are integrated in one embedding. 3) It can be directly trained as a part of the whole model in an end-to-end manner.

\begin{figure}[t]
\centering
\includegraphics[width=0.9\columnwidth]{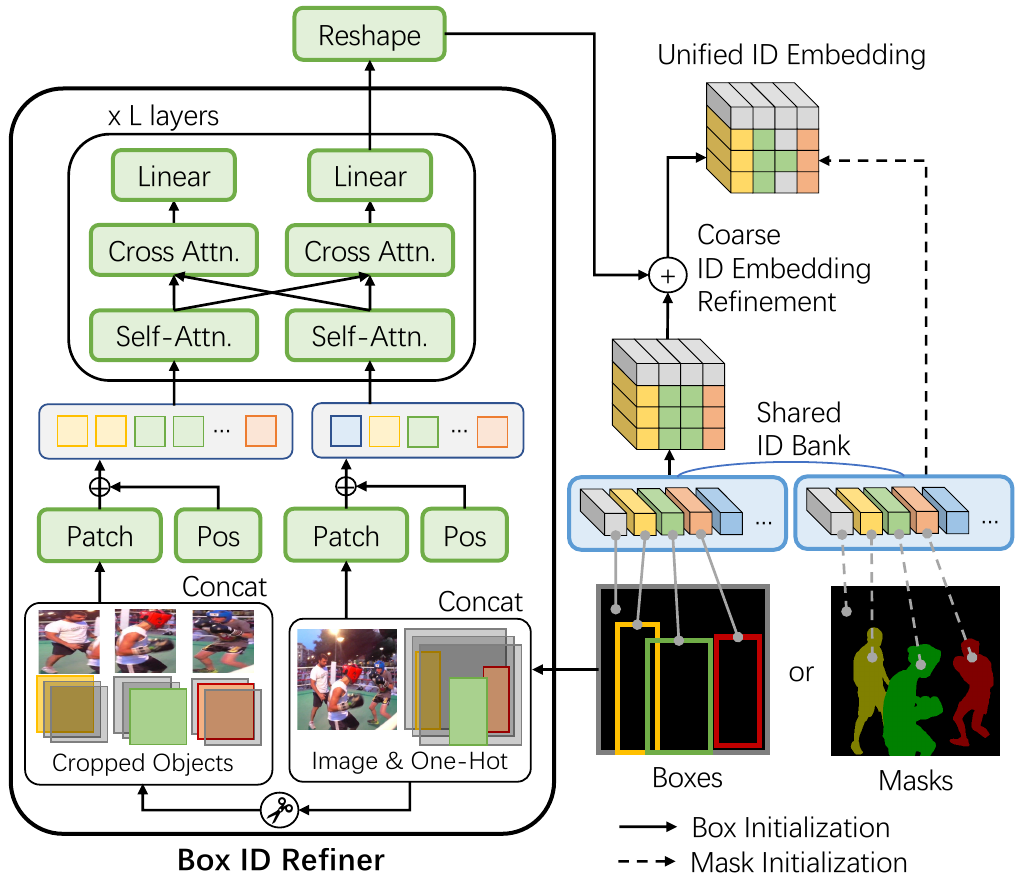}
\caption{Illustration of the unified identification module, which consists of the shared ID bank and the box ID refiner. The procedures of box initialization and mask initialization are both shown. }
\label{fig:UIDM}
\end{figure}

\vspace{-5pt}
\paragraph{Box-Mask Integrated Identification.} The UIDM module consists of two parts, the shared ID bank and the Box ID Refiner (BIDR), as shown in Figure \ref{fig:UIDM}. If the reference is mask representation, the shared ID bank generates an ID embedding directly from the masks, which stores rich object information. If the reference is box representation, the boxes are first converted to box-shaped masks, and the shared ID bank generates a coarse ID embedding. Then the coarse ID embedding is refined by BIDR into a fine ID embedding with more object details, which is unified with the mask reference:
\begin{equation}
E_{id} = 
\begin{cases}
    ID(Y_m), & \text{if $Y_m$ is mask ref.} \\
    ID(Y_m) + R(I_m,Y_m), & \text{if $Y_m$ is box ref.}
\end{cases}
\end{equation}

\vspace{-10pt}
\paragraph{Box ID Refinement.} 
To extract accurate multi-object information from the image according to the boxes, BIDR is designed as a dual path transformer with both self-attention and cross attention layers. One is the image path for global information collection, the other is the object path for local information extraction. First, self-attention is used in both paths. The image path can learn the global information such positions and shapes around whole image including all objects, while the object path can learn detailed information for each object and interactive information between objects such as occlusion. Then, the dual cross attention is used for the information exchange between two paths:
\begin{align}
    &V_O' = Attn(Q_I,K_O,V_O),\ 
    V_I' = Attn(Q_O,K_I,V_I)
\end{align}
where the subscript $I$ is for the image path and $O$ is for the object path. The local information is integrated into the global image path, which helps to generate finer ID embedding. Meanwhile, the global image information also flows to local objects in the boxes, helping to get better object discriminative information, which is favorable in multi-object scenario. After several transformer layers, the refinement embedding can represent more accurate information about the target objects than boxes. The output of BIDR is added with the coarse ID embedding to get a unified fine ID embedding, and then sent to propagation.

\vspace{-5pt}
\paragraph{Auxiliary Mask Reconstruction.} Although the shared ID bank and BIDR in UIDM can be trained in an end-to-end manner, direct learning how to refine a high dimensional embedding may be awkward. We mitigate the problem by employing an extra ID decoder during training. The UIDM acts as an encoder to generate the latent ID embedding, and the ID decoder reconstructs the mask from the latent ID embedding. In this way, the training of UIDM gets easier and more effective. Note that the ID decoder is not used in test.

\subsection{Pinpoint Box Predictor}
\label{sec: box head}

\begin{figure}[t]
\centering
\includegraphics[width=1.0\columnwidth]{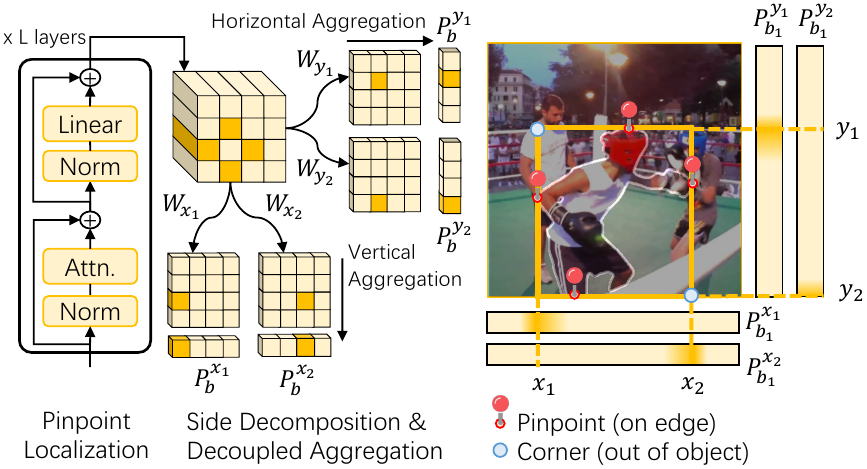}
\caption{Illustration of pinpoint-based box prediction. The pinpoints are first localized by a transformer, and then projected into four probability maps for four sides. Then the probability maps are aggregated horizontally and vertically to extract side-aligned coordinates of pinpoints on each side.}
\label{fig: box head}
\end{figure}

Keypoint-based box head is popular in detection and tracking, such as the corner head \cite{law2018cornernet,yan2021learning,cui2022mixformer} and the center head \cite{duan2019centernet,ye2022joint}. The corner head predicts top left and bottom right points while the center head predicts a box center and box sizes in x and y directions. The corners are exterior points to the objects (even the center does not have to be inside the object). Emphasizing exterior points distracts the learning of accurate object-centric representation, especially when it is shared by both box and mask branch.

\begin{figure}[t]
\centering
\includegraphics[width=1.0\columnwidth]{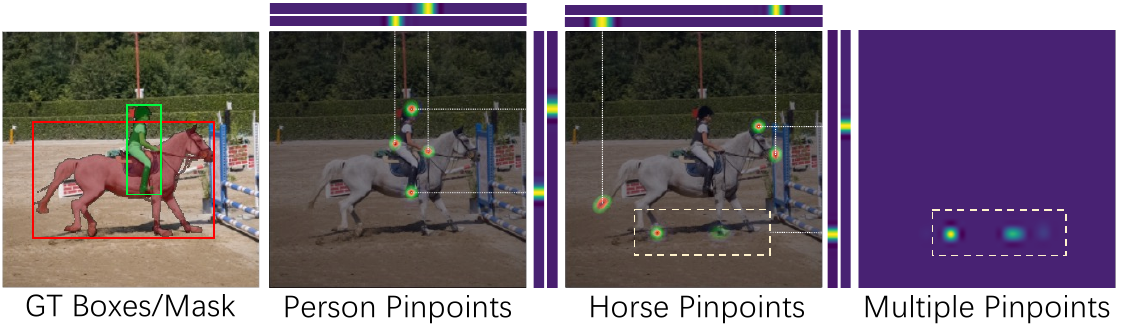}
\caption{Visualization of pinpoint localization and decoupled aggregation. The scene includes two targets, a person and a horse. There are multiple approximate pinpoints on the bottom side of the horse, with each one localized on one hoof.}
\label{fig: pin_viz}
\end{figure}

To make the box predictor harmonious in representation learning with the mask decoder, we propose a pinpoint box head, which determine a box of an object by localizing its pinpoints, illustrated in Figure \ref{fig: box head} and visualized in Figure \ref{fig: pin_viz}. A prior method \cite{zhou2019bottom} has attempted to use pinpoints for object detection, but it still relies on center point for object localization, and requires pinpoint annotations for supervision. However, it is hard to annotate all possible pinpoints for supervision, for example a side-aligned object edge includes infinite pinpoints. To solve this, we propose a decoupled aggregation strategy, which eases the demand for pinpoint annotation for direct supervision.

\vspace{-5pt}
\paragraph{Pinpoint Localization.} Given the propagated embedding $D_t\in R^{HW \times C'}$ as the output from the propagation module, it is fed into several transformer layers with self-attention \cite{vaswani2017attention} to perform pinpoint localization. With the powerful global modeling capacity of attention mechanism, object edge features are recognized. Later, pinpoints can be localized by exploring the extreme positions of the edges.

\vspace{-5pt}
\paragraph{Decoupled Aggregation.} A box has four or more pinpoints according to Property \ref{prop: p1}, while it can be determined by only four parameters, so it's redundant to directly predict all the pinpoints to obtain the box. In addition, there is no pinpoint annotation for direct supervision in our training data. Based on Property \ref{prop: p2}, we propose a decoupled aggregation strategy to predict the box by extracting the side-aligned coordinates from pinpoints on four sides. First, we decompose the pinpoints on the top, bottom, left and right sides from the localization feature maps by convolution layers $W_{x_1,y_1,x_2,y_2}\in R^{C' \times 4M}$ into four score maps. The score maps are normalized by $Softmax$ into probability maps. Then we horizontally aggregate top/bottom pinpoint probability maps, and vertically aggregate left/right pinpoint probability maps into probability vectors $P_b^{x_1,x_2}\in R^{W\times 2M},\ P_b^{y_1,y_2}\in R^{H\times 2M}$:
\begin{align}
P_b^{x_1,x_2} &= \sum_y Softmax(W_{x_1,x_2}\cdot SA(D_t)),\\
P_b^{y_1,y_2} &= \sum_x Softmax(W_{y_1,y_2}\cdot SA(D_t)).
\end{align}
where $SA$ is self-attention. Then the boxes for all $N$ objects are determined by $[\mathbf{x_1},\mathbf{x_2},\mathbf{y_1},\mathbf{y_1}]\in R^{M\times 4}$, which are predicted by $soft-argmax$:
\begin{align}
\mathbf{x_1^T}&=C_x \cdot P_b^{x_1},\ \mathbf{x_2^T}=C_x \cdot P_b^{x_2}\\
\mathbf{y_1^T}&=C_y \cdot P_b^{y_1},\ \mathbf{y_2^T}=C_y \cdot P_b^{y_2}
\end{align}
where $C_x=[0,1,...,W],\ C_y=[0,1,...,H]$ are the coordinate arrangements in x and y directions. Note that we select corresponding $N$ boxes for $N$ objects from $M$.


\vspace{-5pt}
\paragraph{Box-Mask Collaboration.} Although a single mask decoder can predict masks and convert them to bounding boxes, there are two advantages of employing both a mask decoder and a box predictor: \textit{box-mask synergy} in training and \textit{box-mask consistency} during test. We find training two branches together yields better performance than training a single mask branch (Table \ref{tab: box head}), which proves the pinpoint box branch can promote the learning of better edge-aware representation of objects. In addition, the box representation achieves higher precision than the mask representation on VOT benchmarks. During test, we can also calculate the consistency of the two branches by the box IoU, which can be used to indicate the reliability of the results.

\subsection{Loss and Optimization}
The total loss consists of mask loss and box loss. For mask branch, the loss functions we use are Cross Entropy loss and Jaccard loss, and for box branch we use L1 loss and Generalized IoU loss \cite{rezatofighi2019generalized}. For the ID decoder, we only use Cross Entropy loss to supervise the mask reconstruction.

In our framework, both the mask $Y_0^{mask}$ and box $Y_0^{box}$ reference can be used for initialization. The target is to find parameters $\theta$ of the model $f$ for two objectives:
\begin{align}
&\min_{\theta} L(f_\theta(X_0,...,X_{t-1},X_t,Y_0^{mask}),Y_t),\\
&\min_{\theta} L(f_\theta(X_0,...,X_{t-1},X_t,Y_0^{box}),Y_t).
\end{align}
The output of the model $\hat{Y}_t$ and the ground truth $Y_t$ include both masks and boxes, but here we simplify them as a single term. The model needs to achieve the minimal total loss both for box and mask initialization. To solve this, we randomly pick one objective and optimize it in each step and alternate between two objectives. The probability of choosing the box format for initialization is set as $0.3$.

\section{Experiment}
\subsection{Implementation Details}
\paragraph{Model Setting.}  The backbone encoder of MITS is ResNet-50 \cite{he2016deep}. For transformers in UIDM, propagation module and pinpoint box head, we all set 3 layers. The propagation module can be flexible in our framework, and we use the carefully designed gated propagation module proposed in DeAOT \cite{yang2022decoupling} by default. For VOT, we use the box branch for box evaluation by default, and the masks can also be evaluated by taking their bounding boxes.

\vspace{-5pt}
\paragraph{Training.} We use the same training data as RTS \cite{paul2022robust} for fair comparison, including the training sets of LaSOT \cite{fan2019lasot}, GOT-10k \cite{huang2019got}, Youtube-VOS \cite{xu2018youtube} and DAVIS \cite{pont20172017}. For VOT datasets, masks are estimated from boxes by STA \cite{zhao2021generating} that is trained on YouTube-VOS and DAVIS according to RTS. The backbone encoder, propagation module and mask decoder are pre-trained on static images \cite{everingham2010pascal,lin2014microsoft,cheng2014global,shi2015hierarchical,hariharan2011semantic}, following prior VOS methods \cite{yang2021associating,yang2022decoupling}, while UIDM and the box predictor are trained from scratch. During training, we use 4 Nvidia RTX3090 GPUs, and the batch size is 16. The model is trained 100,000 steps with an initial learning rate of $2 \times 10^{-4}$. The learning rate gradually decays to $1 \times 10^{-5}$ in a polynomial manner \cite{yang2020collaborative}. We use only one model to evaluate on all benchmarks, except GOT-10k test set, where only the training set of GOT-10k is used for its one-shot evaluation. 

\subsection{Evaluation on Single Object Tracking}

\begin{table}[]
\centering
\small
\setlength{\tabcolsep}{2pt}
\resizebox{\columnwidth}{!}{%
\begin{tabular}{@{}l|ccccccccc@{}}
\toprule
                                      & \multicolumn{3}{c|}{LaSOT \cite{fan2019lasot}}                                                & \multicolumn{3}{c|}{TrackingNet \cite{muller2018trackingnet}}                                 & \multicolumn{3}{c}{GOT-10k \cite{huang2019got}}                          \\ \midrule
SOT Method                            & AUC                    & P$_{N}$             & \multicolumn{1}{c|}{P}                      & AUC                    & P$_{N}$             & \multicolumn{1}{c|}{P}                      & AO                     & SR$_{0.5}$             & SR$_{0.75}$            \\ \midrule
SiamFC \cite{bertinetto2016fully}     & 33.6                   & 42.0                   & \multicolumn{1}{c|}{33.9}                   & 57.1                   & 66.3                   & \multicolumn{1}{c|}{53.3}                   & 34.8                   & 35.3                   & 9.8                    \\
MDNet \cite{nam2016learning}          & 39.7                   & 46.0                   & \multicolumn{1}{c|}{37.3}                   & 60.6                   & 70.5                   & \multicolumn{1}{c|}{56.5}                   & 29.9                   & 30.3                   & 9.9                    \\
SiamPRN++ \cite{li2019siamrpn++}      & 49.6                   & 56.9                   & \multicolumn{1}{c|}{49.1}                   & 73.3                   & 80.0                   & \multicolumn{1}{c|}{69.4}                   & 51.7                   & 61.6                   & 32.5                   \\
DiMP \cite{bhat2019learning}          & 56.9                   & 65.0                   & \multicolumn{1}{c|}{56.7}                   & 74.0                   & 80.1                   & \multicolumn{1}{c|}{68.7}                   & 61.1                   & 71.7                   & 49.2                   \\
MAMLTrack \cite{wang2020tracking}     & 52.3                   & -                      & \multicolumn{1}{c|}{-}                      & 75.7                   & 82.2                   & \multicolumn{1}{c|}{72.5}                   & -                      & -                      & -                      \\
Ocean \cite{zhang2020ocean}           & 56.0                   & 65.1                   & \multicolumn{1}{c|}{56.6}                   & -                      & -                      & \multicolumn{1}{c|}{-}                      & 61.1                   & 72.1                   & 47.3                   \\
TrDiMP \cite{wang2021transformer}     & 63.9                   & -                      & \multicolumn{1}{c|}{61.4}                   & 78.4                   & 83.3                   & \multicolumn{1}{c|}{73.1}                   & 67.1                   & 77.7                   & 58.3                   \\
TransT \cite{chen2021transformer}     & 64.9                   & 73.8                   & \multicolumn{1}{c|}{69.0}                   & 81.4                   & 86.7                   & \multicolumn{1}{c|}{80.3}                   & 67.1                   & 76.8                   & 60.9                   \\
STARK \cite{yan2021learning}          & 67.1                   & 77.0                   & \multicolumn{1}{c|}{-}                      & 82.0                   & 86.9                   & \multicolumn{1}{c|}{-}                      & 68.8                   & 78.1                   & 64.1                   \\
KeepTrack \cite{mayer2021learning}    & 67.1                   & 77.2                   & \multicolumn{1}{c|}{70.2}                   & -                      & -                      & \multicolumn{1}{c|}{-}                      & -                      & -                      & -                      \\
SBT \cite{xie2022correlation}         & 66.7                   & -                      & \multicolumn{1}{c|}{71.1}                   & 82.2                   & 87.5                   & \multicolumn{1}{c|}{-}                      & 70.4                   & 80.8                   & 64.7                   \\
ToMP \cite{mayer2022transforming}     & 68.5                   & 79.2                   & \multicolumn{1}{c|}{73.5}                   & 81.5                   & 86.4                   & \multicolumn{1}{c|}{78.9}                   & -                      & -                      & -                      \\
MixFormer \cite{cui2022mixformer}     & 70.1                   & 79.9                   & \multicolumn{1}{c|}{76.3}                   & \textcolor{blue}{83.9} & \textcolor{blue}{88.9} & \multicolumn{1}{c|}{83.1}                   & 70.7                   & 80.0                   & 67.8                   \\
AiATrack \cite{gao2022aiatrack}       & 69.0                   & 79.4                   & \multicolumn{1}{c|}{73.8}                   & 82.7                   & 87.8                   & \multicolumn{1}{c|}{80.4}                   & 69.6                   & 80.0                   & 63.2                   \\
OSTrack \cite{ye2022joint}            & \textcolor{blue}{71.1} & \textcolor{red}{81.1}  & \multicolumn{1}{c|}{\textcolor{blue}{77.6}} & \textcolor{red}{83.9}  & 88.5                   & \multicolumn{1}{c|}{\textcolor{blue}{83.2}} & \textcolor{blue}{73.7} & \textcolor{blue}{83.2} & \textcolor{blue}{70.8} \\
SwinTrack \cite{lin2021swintrack}     & 69.6                   & 78.6                   & \multicolumn{1}{c|}{74.1}                   & 82.5                   & 87.0                   & \multicolumn{1}{c|}{80.4}                   & 69.4                   & 78.0                   & 64.3                   \\ \midrule
Unified Method                        &                        &                        &                                             &                        &                        &                                             &                        &                        &                        \\ \midrule
SiamMask \cite{wang2019fast}          & -                      & -                      & \multicolumn{1}{c|}{-}                      & 72.5                   & 77.8                   & \multicolumn{1}{c|}{66.4}                   & 51.4                   & 58.7                   & 36.6                   \\
D3S \cite{lukezic2020d3s}             & -                      & -                      & \multicolumn{1}{c|}{-}                      & 72.8                   & 76.8                   & \multicolumn{1}{c|}{66.4}                   & 59.7                   & 67.6                   & 46.2                   \\
SiamR-CNN \cite{voigtlaender2020siam} & 64.8                   & 72.2                   & \multicolumn{1}{c|}{-}                      & 81.2                   & 85.4                   & \multicolumn{1}{c|}{80.0}                   & 64.9                   & 72.8                   & 59.7                   \\
Unicorn \cite{yan2022towards}         & 68.5                   & 76.6                   & \multicolumn{1}{c|}{74.1}                   & 83.0                   & 86.4                   & \multicolumn{1}{c|}{82.2}                   & -                      & -                      & -                      \\
RTS \cite{paul2022robust}             & 69.7                   & 76.2                   & \multicolumn{1}{c|}{73.7}                   & 81.6                   & 86.0                   & \multicolumn{1}{c|}{79.4}                   & -                      & -                      & -                      \\ \midrule
\textbf{MITS (Ours)}                                  & \textcolor{red}{72.0}  & \textcolor{blue}{80.1} & \multicolumn{1}{c|}{\textcolor{red}{78.5}}  & 83.4                   & \textcolor{red}{88.9}  & \multicolumn{1}{c|}{\textcolor{red}{84.6}}  & \textcolor{red}{80.4}  & \textcolor{red}{89.8}  & \textcolor{red}{75.8}  \\ \bottomrule
\end{tabular}%
}
\caption{Evaluation results on SOT benchmarks. There are two groups of methods 1) SOT only methods, 2) unified tracking and segmentation methods. Specially, one-shot setting is followed on GOT-10k test. The best two results are shown in \textcolor{red}{red} and \textcolor{blue}{blue}.}
\label{tab:SOT}
\end{table}

\begin{table}[]
\centering
\small
\setlength{\tabcolsep}{2pt}
\resizebox{\columnwidth}{!}{%
\begin{tabular}{@{}lc|cccccccc@{}}
\toprule
                                      & \multicolumn{1}{l|}{} & \multicolumn{5}{c|}{YouTube-VOS 19 Val \cite{xu2018youtube}}                                                                                    & \multicolumn{3}{c}{DAVIS 17 Val \cite{pont20172017}}                         \\ \midrule
VOS Method                            & Init.                 & $\mathcal{G}$          & \multicolumn{2}{c}{$\mathcal{J/F}_{seen}$}      & \multicolumn{2}{c|}{$\mathcal{J/F}_{unseen}$}                        & $\mathcal{J}\&\mathcal{F}$ & $\mathcal{J}$          & $\mathcal{F}$          \\ \midrule
STM \cite{oh2019video}                & M                     & 79.2                   & 79.6                   & 73.0                   & 86.3                   & \multicolumn{1}{c|}{80.6}                   & 81.8                       & 79.2                   & 84.3                   \\
CFBI \cite{yang2020collaborative}     & M                     & 81.0                   & 80.6                   & 85.1                   & 75.2                   & \multicolumn{1}{c|}{83.0}                   & 81.9                       & 79.1                   & 84.6                   \\
SST \cite{duke2021sstvos}             & M                     & 81.8                   & 80.9                   & -                      & 76.7                   & \multicolumn{1}{c|}{-}                      & -                          & -                      & -                      \\
HMMN \cite{seong2021hierarchical}     & M                     & 82.5                   & 81.7                   & 86.1                   & 77.3                   & \multicolumn{1}{c|}{85.0}                   & 84.7                       & 81.9                   & 87.5                   \\
CFBI+ \cite{yang2021collaborative}    & M                     & 82.6                   & 81.7                   & 86.2                   & 77.1                   & \multicolumn{1}{c|}{85.2}                   & -                          & -                      & -                      \\
JOINT \cite{mao2021joint}             & M                     & 82.8                   & 80.8                   & 84.8                   & 79.0                   & \multicolumn{1}{c|}{86.6}                   & 83.5                       & 80.8                   & 86.2                   \\
STCN \cite{cheng2021rethinking}       & M                     & 82.7                   & 81.1                   & 85.4                   & 78.2                   & \multicolumn{1}{c|}{85.9}                   & \textcolor{blue}{85.4}     & \textcolor{blue}{82.2} & \textcolor{blue}{88.6} \\
AOT \cite{yang2021associating}        & M                     & 85.3                   & 83.9                   & 88.8                   & 79.9                   & \multicolumn{1}{c|}{88.5}                   & 84.9                       & 82.3                   & 87.5                   \\
XMem \cite{cheng2022xmem}             & M                     & 85.5                   & 84.3                   & 88.6                   & 80.3                   & \multicolumn{1}{c|}{88.6}                   & \textcolor{red}{86.2}      & \textcolor{red}{82.9}  & \textcolor{red}{89.5}  \\
DeAOT \cite{yang2022decoupling}       & M                     & \textcolor{blue}{85.9} & \textcolor{blue}{84.6} & \textcolor{blue}{89.4} & \textcolor{red}{80.8}  & \multicolumn{1}{c|}{\textcolor{red}{88.9}}  & 85.2                       & 82.2                   & 88.2                   \\ \midrule
Unified Method                        & \multicolumn{1}{l|}{} &                        &                        &                        &                        &                                             &                            &                        &                        \\ \midrule
D3S \cite{lukezic2020d3s}             & M                     & -                      & -                      & -                      & -                      & \multicolumn{1}{c|}{-}                      & 60.8                       & 57.8                   & 63.8                   \\
SiamR-CNN \cite{voigtlaender2020siam} & M                     & -                      & -                      & -                      & -                      & \multicolumn{1}{c|}{-}                      & 74.8                       & 69.3                   & 80.2                   \\
Unicorn \cite{yan2022towards}         & M                     & -                      & -                      & -                      & -                      & \multicolumn{1}{c|}{-}                      & 69.2                       & 65.2                   & 73.2                   \\
RTS \cite{paul2022robust}             & M                     & 79.7                   & 77.9                   & 75.4                   & 82.0                   & \multicolumn{1}{c|}{83.3}                   & 80.2                       & 77.9                   & 82.6                   \\ \midrule
\textbf{MITS (Ours)}                                  & M                     & \textcolor{red}{85.9}  & \textcolor{red}{84.9}  & \textcolor{red}{89.7}  & \textcolor{blue}{80.4} & \multicolumn{1}{c|}{\textcolor{blue}{88.7}} & 84.9                       & 82.0                   & 87.7                   \\ \midrule\midrule
Box Init.                             &                       &                        &                        &                        &                        &                                             &                            &                        &                        \\ \midrule
SiamMask \cite{wang2019fast}          & B                     & 52.8                   & 60.2                   & 45.1                   & 58.2                   & \multicolumn{1}{c|}{47.7}                   & 56.4                       & 54.3                   & 58.5                   \\
SiamR-CNN \cite{voigtlaender2020siam} & B                     & 67.3                   & 68.1                   & 61.5                   & 70.8                   & \multicolumn{1}{c|}{68.8}                   & 70.6                       & 66.1                   & 75.0                   \\
DeAOT \cite{yang2022decoupling}       & B                     & 68.2                   & 67.9                   & \textcolor{blue}{69.6} & 64.0                   & \multicolumn{1}{c|}{71.3}                   & 64.0                       & 61.6                   & 66.4                   \\
RTS \cite{paul2022robust}             & B                     & \textcolor{blue}{70.8} & \textcolor{blue}{71.1} & 65.2                   & \textcolor{blue}{74.0} & \multicolumn{1}{c|}{\textcolor{blue}{72.8}}                   & \textcolor{blue}{72.6}     & \textcolor{blue}{69.4} & \textcolor{blue}{75.8} \\ \midrule
\textbf{MITS (Ours)}                                  & B                     & \textcolor{red}{81.8}  & \textcolor{red}{81.9}  & \textcolor{red}{86.2}  & \textcolor{red}{75.2}  & \multicolumn{1}{c|}{\textcolor{red}{83.7}}  & \textcolor{red}{81.1}      & \textcolor{red}{77.5}  & \textcolor{red}{84.7}  \\ \bottomrule
\end{tabular}%
}
\caption{Evaluation results on VOS benchmarks. There are two groups of methods 1) VOS only methods, 2) unified tracking and segmentation methods. Different initialization formats, mask (M) or box (B), are compared separately. The best two results are shown in \textcolor{red}{red} and \textcolor{blue}{blue}.}
\label{tab:VOS}
\end{table}

\paragraph{LaSOT \& TrackingNet.} LaSOT \cite{fan2019lasot} is a long-term tracking benchmark consisting of 280 videos in test set. The videos are at 30 FPS and have an average length around 2.5k frames. TrackingNet \cite{muller2018trackingnet} is a short-term tracking benchmark, which contains 511 videos for test, and the average length is around 0.45k frames. The evaluation metrics are Success Rate (or AUC, Area Under Curve), Precision (P) and Normalized Precision (P$_N$).

\begin{figure*}[t]
\centering
\includegraphics[width=0.95\textwidth]{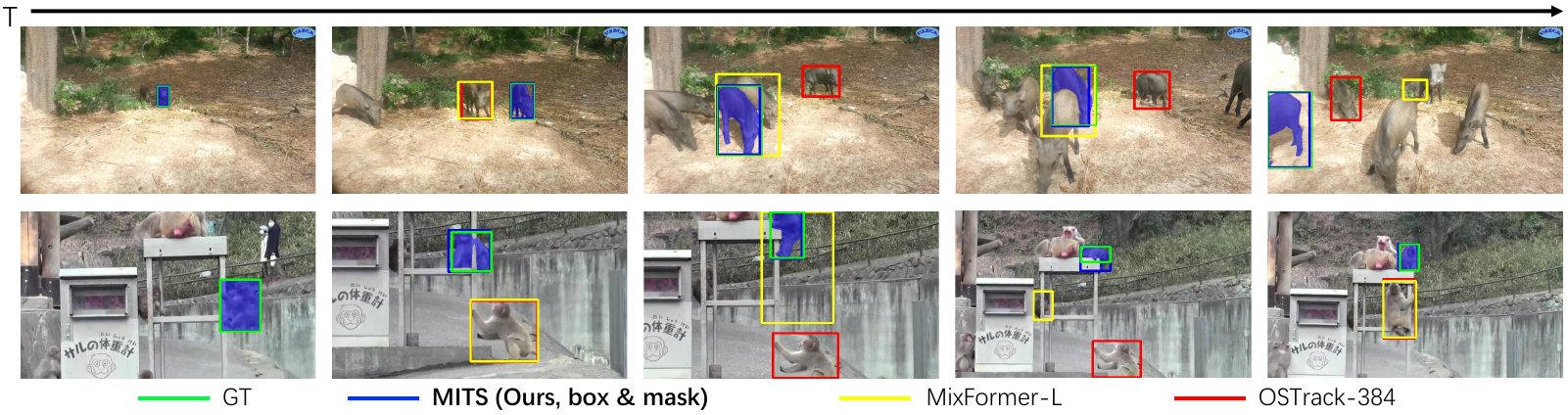}
\caption{Qualitative results of MITS on VOT, compared with SOTA SOT methods MixFormer \cite{cui2022mixformer} and OSTrack \cite{ye2022joint}.}
\label{fig: qua_VOT2}
\end{figure*}

We select very strong SOT methods for comparison in Table \ref{tab:SOT}, such as MixFormer \cite{cui2022mixformer} and OSTrack \cite{ye2022joint}, where large-scale transformers \cite{dosovitskiy2020image,wu2021cvt} pretrained on ImageNet-22k \cite{deng2009imagenet} or by MAE \cite{he2022masked} are used. Despite MITS only uses a ResNet-50 \cite{he2016deep} backbone, it still achieves highest Success Rate/Precision on LaSOT and highest Normalized Precision/Precision on TrackingNet. Although MITS searches objects in the whole image, it runs at 20 FPS, which is comparable with MixFormer. MITS also significantly improve the performance of unified methods such as Unicorn \cite{ma2022unified}, by 3.5\% Success Rate and 4.4\% Precision on LaSOT. MITS achieves impressive Precision owing to the accurate localization and robust decoupled prediction of the pinpoint head. Qualitative results are shown in Figure \ref{fig: qua_VOT2}. We select video sequences which have multiple similar objects, and MITS tracks the target more robustly than other SOT methods, benefited from the multi-object framework.

\vspace{-5pt}
\paragraph{GOT-10k.} GOT-10k \cite{huang2019got} is a large-scale short-term tracking benchmark, offering 9.34k videos for training, 180 videos for validation and 420 videos for test. We follow the \textit{one-shot protocol} to verify the generalization ability of MITS on tracking, which requires GOT-10k training only to avoid biased results. MITS achieves absolutely superior performance on GOT-10k, with around 6\% improvement over prior SOTA SOT method OSTrack \cite{ye2022joint}. The improvement can be attributed to two aspects. First, masks with detailed object information are used in training. UIDM learns to extract detailed spatial information from boxes, which leads to accurate initialization. Second, MITS propagates the detailed spatial information from multiple memory frames with a long temporal coverage, and the prediction is on the whole image. Considering the videos are at 10FPS, the impressive results suggest that MITS is capable of robustly tracking objects in low frame rate videos, where motion is faster than that in high frame rate videos. SOT methods usually have a strong temporal assumption (searching in a local area) and may fail to capture fast moving objects.


\subsection{Evaluation on Video Object Segmentation}

\begin{figure*}[t]
\centering
\includegraphics[width=0.9\textwidth]{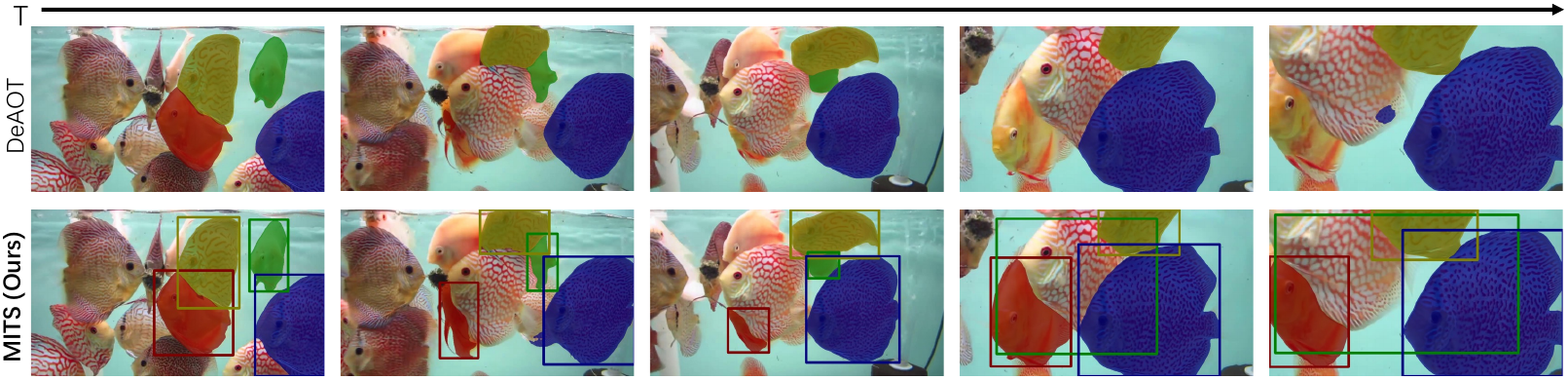}
\caption{Qualitative results of MITS on VOS, compared with SOTA VOS method DeAOT \cite{yang2022decoupling}. For MITS, predictions from box predictor and mask decoder are both visualized.}
\label{fig: qua_VOS2}
\end{figure*}

\paragraph{YouTube-VOS 2019 \& DAVIS 2017.} DAVIS \cite{pont20172017} is a small-scale benchmark for VOS, which contains 30 videos with high quality annotation masks in 2017 validation set. YouTube-VOS \cite{xu2018youtube} is a large-scale benchmark for VOS, including 507 videos in 2019 validation set with 65/26 seen/unseen object categories. The Region Jaccard $\mathcal{J}$ and the Boundary F measure $\mathcal{F}$ are used as metrics.

It can be found in Table \ref{tab:VOS} that the performance of most prior unified methods is not satisfactory compared with VOS only methods, and MITS achieves SOTA performance on YouTube-VOS. MITS also performs better than all previous unified methods on DAVIS 2017 validation set and also competitive with recent VOS methods \cite{yang2021associating}. Qualitative results are shown in Figure \ref{fig: qua_VOS2}.

\vspace{-5pt}
\paragraph{Box Initialization.} Following prior work \cite{voigtlaender2020siam,paul2022robust}, we also tried using box initialization on VOS benchmarks. MITS with box initialization even outperforms prior unified methods with mask initialization on both YouTube-VOS and DAVIS. Besides, although DeAOT performs well on the VOS task, it relies heavily on fine masks for good results. If only boxes are available, it will suffer from sever performance degradation by around -20\%. Differently, MITS only drops a small fraction from mask to box initialization, which shows the strong compatibility of UIDM.


\vspace{-5pt}
\section{Ablation Study}
\subsection{Unified Identification Module}
Ablation results of UIDM are listed in Table \ref{tab:UIDM}. First, we compare MITS with DeAOT \cite{yang2022decoupling} on the tracking benchmark LaSOT \cite{fan2019lasot}. Since DeAOT is a VOS method, we use off-the-shelf Alpha-Refine \cite{yan2021alpha} model to generate masks from boxes for initialization. Compared with the separate models, our unified framework with UIDM achieves better performance. Second, we provide some ablation results of the box ID refiner (BIDR) in UIDM. During training, masks are reconstructed from the refined ID embedding (Mask Recon.), which leads to 0.5\% Success Rate/1.7\% Precision improvement. The dual cross attention (Dual CA) is used in BIDR to effectively exchange information between the global image and local cropped boxes. If only a single cross attention for the image path in each layer of BIDR is used, the Success Rate drops 1.1\% and Precision drops 2\%. The bidirectional information exchange is crucial in extracting detailed spatial information from boxes.


\begin{table}[]
\centering
\small
\renewcommand{\arraystretch}{0.8}
\begin{tabular}{@{}ll|ccc@{}}
\toprule
                                &                                 & \multicolumn{3}{c}{LaSOT \cite{fan2019lasot}}                     \\
Method                          & Box-Mask Module                 & AUC           & P$_{N}$    & P             \\ \midrule
DeAOT \cite{yang2022decoupling} & None                            & 65.9          & 70.7          & 68.4          \\
DeAOT \cite{yang2022decoupling} & AlphaRefine \cite{yan2021alpha} & 67.5          & 72.6          & 71.1          \\
DeAOT$^{\dag}$ \cite{yang2022decoupling} & AlphaRefine \cite{yan2021alpha} & 71.1          & 79.4          & 77.9          \\\midrule
MITS                            & UIDM                            & \textbf{72.0} & \textbf{80.1} & \textbf{78.5} \\
MITS                            & UIDM, no Mask Recon.            & 71.5          & 79.0          & 76.8          \\
MITS                            & UIDM, no Dual CA                & 70.9          & 78.2          & 76.4          \\ \bottomrule
\end{tabular}%
\caption{Ablation of the unified identification module. We also compare MITS with SOTA VOS method DeAOT \cite{yang2022decoupling} with extra model AlphaRefine \cite{yan2021alpha} on the VOT task. $\dag$: DeAOT re-trained with the same data as MITS.
\vspace{-5pt}}
\label{tab:UIDM}
\end{table}

\vspace{-5pt}
\subsection{Pinpoint Box Predictor}
Ablation results of the box predictor is in Table \ref{tab: box head}. We tried corner head, and other variants of the pinpoint head. Compared with a single mask branch, the mask branch is promoted by the box branch during training. Except explicitly predicting the probability maps of pinpoints, another choice is to localize the pinpoints implicitly in high dimension feature maps. The decoupled aggregation on probability maps is replaced by decoupled pooling on localized feature maps. The feature maps are pooled to vectors and then projected to probability vectors. The implicit pinpoint localization performs worse than explicit version on Success Rate. In addition, we also tried aligning box predictor and mask decoder with the similar FPN \cite{lin2017feature}, but the result is not as good as transformer \cite{vaswani2017attention}.

\begin{table}[]
\centering
\small
\setlength{\tabcolsep}{2pt}
\renewcommand{\arraystretch}{0.8}
\begin{tabular}{@{}ll|ccc|ccc@{}}
\toprule
                                        &                   & \multicolumn{3}{c|}{Box Branch}               & \multicolumn{3}{c}{Mask Branch}               \\
Localization                            & Predict.          & AUC           & P$_{N}$       & P             & AUC           & P$_{N}$       & P             \\ \midrule
None                                    & None              & -             & -             & -             & 71.5          & 78.8          & 77.2          \\
CornerNet \cite{law2018cornernet}       & Corner            & 69.1          & 79.5          & 75.9          & 71.7          & 78.5          & 77.6          \\ \midrule
FPN \cite{lin2017feature}               & Pinpoint$^{\ast}$ & 70.6          & 78.9          & 77.3          & 71.2          & 78.8          & 78.4          \\
Transformer \cite{vaswani2017attention}  & Pinpoint$^{\ast}$ & 71.1          & 79.9          & 78.4          & 72.0          & 79.9          & \textbf{78.4} \\
Transformer \cite{vaswani2017attention} & Pinpoint          & \textbf{72.0} & \textbf{80.1} & \textbf{78.5} & \textbf{72.3} & \textbf{79.9} & 78.2          \\ \bottomrule
\end{tabular}%
\caption{Ablation results of the box predictor on LaSOT \cite{fan2019lasot}. We tried corner head, and other pinpoint head variants like implicit pinpoint localization (marked by $\ast$).}
\label{tab: box head}
\end{table}

\begin{table}[]
\centering
\setlength{\tabcolsep}{3pt}
\renewcommand{\arraystretch}{0.8}
\resizebox{\columnwidth}{!}{%
\begin{tabular}{@{}llc|ccc|c@{}}
\toprule
                                          &                                 & \multicolumn{1}{l|}{} & \multicolumn{3}{c|}{LaSOT \cite{fan2019lasot}} & \multicolumn{1}{l}{}    \\
Backbone                                  & PM                              & Layer                 & AUC            & P$_{N}$    & P             & \multicolumn{1}{l}{FPS} \\ \midrule
ResNet-50 \cite{he2016deep}               & GPM \cite{yang2022decoupling}   & 3                     & \textbf{72.0}  & \textbf{80.1} & \textbf{78.5} & 20                      \\
ResNet-50 \cite{he2016deep}               & LSTT \cite{yang2021associating} & 3                     & 69.3           & 76.3          & 74.5          & 13                      \\
MobileNetV2 \cite{sandler2018mobilenetv2} & GPM \cite{yang2022decoupling}   & 2                     & 67.8           & 74.6          & 72.6          & \textbf{27}             \\ \bottomrule
\end{tabular}%
}
\caption{Results of different propagation modules (PM) and backbones in our framework.}
\label{tab:prop}
\end{table}

\subsection{Propagation Module}
The propagation module can be flexible in our framework, and a general form has been shown in Equation \ref{eq:prop}. In practice, we use the carefully designed Gated Propagation Module (GPM) proposed in DeAOT \cite{yang2022decoupling} by default, and the number of GPM layer is 3. We also tried other variants including 2-layer GPM or 3-layer Long Short-Term Transformer (LSTT) proposed in AOT \cite{yang2021associating}. Readers can refer to AOT \cite{yang2021associating} and DeAOT \cite{yang2022decoupling} for more detailed structures of the propagation modules. As the experimental results show in Table \ref{tab:prop}, GPM performs better than LSTT on both efficiency and performance. Fewer layers of propagation module and a more lightweight backbone lead to the trade-off between efficiency and performance.

\subsection{Training Data}
There are two options for training data: (A) VOS box\&mask + VOT box\&pseudo-mask, (B) VOS box\&mask + VOT box. (A) is taken as default and we directly borrow the generated pseudo-masks for VOT data from prior work \cite{paul2022robust}. While due to the strong compatibility of MITS, (B) is also originally supported for training. Table \ref{tab: box_train} shows that MITS trained with (B) can still achieve strong results. We use different dataset repeats to balance different sources.


\begin{table}[]
\centering
\setlength{\tabcolsep}{1.5pt}
\renewcommand{\arraystretch}{0.8}
\resizebox{\columnwidth}{!}{%
\begin{tabular}{@{}lcc|cc|c|c|c@{}}
\toprule
\multicolumn{3}{c|}{VOT data} & \multicolumn{2}{c|}{TrackingNet \cite{muller2018trackingnet}} & LaSOT \cite{fan2019lasot} & \multicolumn{1}{l|}{GOT10k \cite{huang2019got}} & YTB19\cite{xu2018youtube} \\
        & Mask          & Box           & AUC                        & P$_{N}$                       & AUC                       & AO                                              & $\mathcal{G}$             \\ \midrule
(A)     & \ding{51}     & \ding{51}     & 83.4                       & 88.9                             & 72.0                      & 78.5$^\ast$                                     & 85.9                      \\
(B)     & \ding{55}     & \ding{51}     & 82.8                       & 87.9                             & 70.7                      & 78.0$^\ast$                                     & 85.7                      \\ \bottomrule
\end{tabular}%
}
\caption{Comparison between different training data strategies. Pseudo-masks for VOT data is used in (A) but not in (B), with the mask loss disabled for VOT data. $\ast$: results for GOT10k are trained with all datasets rather than one-shot protocol.}
\label{tab: box_train}
\end{table}

\begin{table}[]
\centering
\small
\renewcommand{\arraystretch}{0.8}
\begin{tabular}{@{}l|ccc|ccc@{}}
\toprule
              & \multicolumn{3}{c|}{LaSOT \cite{fan2019lasot}} & \multicolumn{3}{c}{TrackingNet \cite{muller2018trackingnet}} \\
Branch        & AUC            & P$_{N}$       & P             & AUC                & P$_{N}$            & P                  \\ \midrule
Box Predictor & 72.0           & \textbf{80.1} & \textbf{78.5} & 83.4               & \textbf{88.9}      & \textbf{84.6}      \\
Mask Decoder  & \textbf{72.3}  & 79.9          & 78.2          & \textbf{83.5}      & 88.5               & 84.0               \\ \bottomrule
\end{tabular}%
\caption{Comparison between box prediction and mask prediction on SOT benchmarks.}
\label{tab: branch}
\end{table}

\subsection{Box and Mask Prediction}
In our framework, we employ two branches, a mask decoder and a box predictor for joint training and prediction. Training with two branches yields better results than one branch. If the model is trained without pseudo-masks of VOT data, the box branch would perform better than mask branch on VOT benchmarks. If trained with full data, the two branch will reach similar performance, but they also have different prediction strategies. Previous work \cite{paul2022robust} has shown mask representation is more robust than box representation. However, we find the box representation has higher Precision while the mask representation has higher Success Rate, as shown in Table \ref{tab: branch}. 

Another difference between two branches is on the prediction when the target object is absent. SOT benchmarks like LaSOT \cite{fan2019lasot} ignore the frames when the object is absent (fully occluded or out of view), while VOS benchmarks like YouTube-VOS \cite{xu2018youtube} requires to predict an empty mask with all background. Some SOT methods \cite{cui2022mixformer,ye2022joint} predicts boxes for all frames without considering whether the object is absent. As a unified framework, MITS can infer whether an object is absent with the help of the mask prediction, as shown in Figure \ref{fig: qua_VOS2}.



\vspace{-5pt}
\section{Conclusion}
In this paper, we propose a multi-object box-mask integrated framework for unified visual object tracking and segmentation. The proposed unified identification module shows strong compatibility on both box and mask initialization. Refining the ID embedding from boxes helps the model to capture more detailed spatial information about target objects. Besides, the pinpoint box predictor is proposed for accurate box prediction, considering the joint learning of both box and mask branches. The pinpoint head not only facilitates the target-oriented representation learning, but also enables more flexible training strategies for MITS. As a result, our framework achieves SOTA performance on both VOT and VOS benchmarks.

\noindent\textbf{Acknowledgements.} This work was supported by the Fundamental Research Funds for the Central Universities~(No.~226-2023-00048, No.~226-2023-00051).

{\small
\bibliographystyle{ieee_fullname}
\bibliography{egbib}

\begin{thebibliography}{100}\itemsep=-1pt

\bibitem{andriluka2018posetrack}
Mykhaylo Andriluka, Umar Iqbal, Eldar Insafutdinov, Leonid Pishchulin, Anton Milan, Juergen Gall, and Bernt Schiele.
\newblock Posetrack: A benchmark for human pose estimation and tracking.
\newblock In {\em Proceedings of the IEEE conference on computer vision and pattern recognition}, pages 5167--5176, 2018.

\bibitem{bertinetto2016fully}
Luca Bertinetto, Jack Valmadre, Joao~F Henriques, Andrea Vedaldi, and Philip~HS Torr.
\newblock Fully-convolutional siamese networks for object tracking.
\newblock In {\em Computer Vision--ECCV 2016 Workshops: Amsterdam, The Netherlands, October 8-10 and 15-16, 2016, Proceedings, Part II 14}, pages 850--865. Springer, 2016.

\bibitem{bhat2019learning}
Goutam Bhat, Martin Danelljan, Luc~Van Gool, and Radu Timofte.
\newblock Learning discriminative model prediction for tracking.
\newblock In {\em Proceedings of the IEEE/CVF international conference on computer vision}, pages 6182--6191, 2019.

\bibitem{bhat2020learning}
Goutam Bhat, Felix~J{\"a}remo Lawin, Martin Danelljan, Andreas Robinson, Michael Felsberg, Luc~Van Gool, and Radu Timofte.
\newblock Learning what to learn for video object segmentation.
\newblock In {\em European Conference on Computer Vision}, pages 777--794. Springer, 2020.

\bibitem{bolme2010visual}
David~S Bolme, J~Ross Beveridge, Bruce~A Draper, and Yui~Man Lui.
\newblock Visual object tracking using adaptive correlation filters.
\newblock In {\em 2010 IEEE computer society conference on computer vision and pattern recognition}, pages 2544--2550. IEEE, 2010.

\bibitem{caelles2017one}
Sergi Caelles, Kevis-Kokitsi Maninis, Jordi Pont-Tuset, Laura Leal-Taix{\'e}, Daniel Cremers, and Luc Van~Gool.
\newblock One-shot video object segmentation.
\newblock In {\em Proceedings of the IEEE conference on computer vision and pattern recognition}, pages 221--230, 2017.

\bibitem{chen2022backbone}
Boyu Chen, Peixia Li, Lei Bai, Lei Qiao, Qiuhong Shen, Bo Li, Weihao Gan, Wei Wu, and Wanli Ouyang.
\newblock Backbone is all your need: a simplified architecture for visual object tracking.
\newblock In {\em Computer Vision--ECCV 2022: 17th European Conference, Tel Aviv, Israel, October 23--27, 2022, Proceedings, Part XXII}, pages 375--392. Springer, 2022.

\bibitem{chen2021transformer}
Xin Chen, Bin Yan, Jiawen Zhu, Dong Wang, Xiaoyun Yang, and Huchuan Lu.
\newblock Transformer tracking.
\newblock In {\em Proceedings of the IEEE/CVF conference on computer vision and pattern recognition}, pages 8126--8135, 2021.

\bibitem{cheng2022xmem}
Ho~Kei Cheng and Alexander~G Schwing.
\newblock Xmem: Long-term video object segmentation with an atkinson-shiffrin memory model.
\newblock In {\em Computer Vision--ECCV 2022: 17th European Conference, Tel Aviv, Israel, October 23--27, 2022, Proceedings, Part XXVIII}, pages 640--658. Springer, 2022.

\bibitem{cheng2021rethinking}
Ho~Kei Cheng, Yu-Wing Tai, and Chi-Keung Tang.
\newblock Rethinking space-time networks with improved memory coverage for efficient video object segmentation.
\newblock {\em Advances in Neural Information Processing Systems}, 34:11781--11794, 2021.

\bibitem{cheng2014global}
Ming-Ming Cheng, Niloy~J Mitra, Xiaolei Huang, Philip~HS Torr, and Shi-Min Hu.
\newblock Global contrast based salient region detection.
\newblock {\em IEEE transactions on pattern analysis and machine intelligence}, 37(3):569--582, 2014.

\bibitem{samtrack}
Yangming Cheng, Liulei Li, Yuanyou Xu, Xiaodi Li, Zongxin Yang, Wenguan Wang, and Yi Yang.
\newblock Segment and track anything.
\newblock {\em arXiv preprint arXiv:2305.06558}, 2023.

\bibitem{cui2022mixformer}
Yutao Cui, Cheng Jiang, Limin Wang, and Gangshan Wu.
\newblock Mixformer: End-to-end tracking with iterative mixed attention.
\newblock In {\em Proceedings of the IEEE/CVF Conference on Computer Vision and Pattern Recognition}, pages 13608--13618, 2022.

\bibitem{danelljan2017eco}
Martin Danelljan, Goutam Bhat, Fahad Shahbaz~Khan, and Michael Felsberg.
\newblock Eco: Efficient convolution operators for tracking.
\newblock In {\em Proceedings of the IEEE conference on computer vision and pattern recognition}, pages 6638--6646, 2017.

\bibitem{danelljan2020probabilistic}
Martin Danelljan, Luc~Van Gool, and Radu Timofte.
\newblock Probabilistic regression for visual tracking.
\newblock In {\em Proceedings of the IEEE/CVF conference on computer vision and pattern recognition}, pages 7183--7192, 2020.

\bibitem{danelljan2016beyond}
Martin Danelljan, Andreas Robinson, Fahad Shahbaz~Khan, and Michael Felsberg.
\newblock Beyond correlation filters: Learning continuous convolution operators for visual tracking.
\newblock In {\em Computer Vision--ECCV 2016: 14th European Conference, Amsterdam, The Netherlands, October 11-14, 2016, Proceedings, Part V 14}, pages 472--488. Springer, 2016.

\bibitem{deng2009imagenet}
Jia Deng, Wei Dong, Richard Socher, Li-Jia Li, Kai Li, and Li Fei-Fei.
\newblock Imagenet: A large-scale hierarchical image database.
\newblock In {\em 2009 IEEE conference on computer vision and pattern recognition}, pages 248--255. Ieee, 2009.

\bibitem{dosovitskiy2020image}
Alexey Dosovitskiy, Lucas Beyer, Alexander Kolesnikov, Dirk Weissenborn, Xiaohua Zhai, Thomas Unterthiner, Mostafa Dehghani, Matthias Minderer, Georg Heigold, Sylvain Gelly, et~al.
\newblock An image is worth 16x16 words: Transformers for image recognition at scale.
\newblock {\em arXiv preprint arXiv:2010.11929}, 2020.

\bibitem{du2020correlation}
Fei Du, Peng Liu, Wei Zhao, and Xianglong Tang.
\newblock Correlation-guided attention for corner detection based visual tracking.
\newblock In {\em Proceedings of the IEEE/CVF Conference on Computer Vision and Pattern Recognition}, pages 6836--6845, 2020.

\bibitem{duan2019centernet}
Kaiwen Duan, Song Bai, Lingxi Xie, Honggang Qi, Qingming Huang, and Qi Tian.
\newblock Centernet: Keypoint triplets for object detection.
\newblock In {\em Proceedings of the IEEE/CVF international conference on computer vision}, pages 6569--6578, 2019.

\bibitem{duke2021sstvos}
Brendan Duke, Abdalla Ahmed, Christian Wolf, Parham Aarabi, and Graham~W Taylor.
\newblock Sstvos: Sparse spatiotemporal transformers for video object segmentation.
\newblock In {\em Proceedings of the IEEE/CVF Conference on Computer Vision and Pattern Recognition}, pages 5912--5921, 2021.

\bibitem{everingham2010pascal}
Mark Everingham, Luc Van~Gool, Christopher~KI Williams, John Winn, and Andrew Zisserman.
\newblock The pascal visual object classes (voc) challenge.
\newblock {\em International journal of computer vision}, 88(2):303--338, 2010.

\bibitem{fan2019lasot}
Heng Fan, Liting Lin, Fan Yang, Peng Chu, Ge Deng, Sijia Yu, Hexin Bai, Yong Xu, Chunyuan Liao, and Haibin Ling.
\newblock Lasot: A high-quality benchmark for large-scale single object tracking.
\newblock In {\em Proceedings of the IEEE/CVF conference on computer vision and pattern recognition}, pages 5374--5383, 2019.

\bibitem{gao2022aiatrack}
Shenyuan Gao, Chunluan Zhou, Chao Ma, Xinggang Wang, and Junsong Yuan.
\newblock Aiatrack: Attention in attention for transformer visual tracking.
\newblock In {\em Computer Vision--ECCV 2022: 17th European Conference, Tel Aviv, Israel, October 23--27, 2022, Proceedings, Part XXII}, pages 146--164. Springer, 2022.

\bibitem{girshick2015fast}
Ross Girshick.
\newblock Fast r-cnn.
\newblock In {\em Proceedings of the IEEE international conference on computer vision}, pages 1440--1448, 2015.

\bibitem{guo2020siamcar}
Dongyan Guo, Jun Wang, Ying Cui, Zhenhua Wang, and Shengyong Chen.
\newblock Siamcar: Siamese fully convolutional classification and regression for visual tracking.
\newblock In {\em Proceedings of the IEEE/CVF conference on computer vision and pattern recognition}, pages 6269--6277, 2020.

\bibitem{hariharan2011semantic}
Bharath Hariharan, Pablo Arbel{\'a}ez, Lubomir Bourdev, Subhransu Maji, and Jitendra Malik.
\newblock Semantic contours from inverse detectors.
\newblock In {\em 2011 international conference on computer vision}, pages 991--998. IEEE, 2011.

\bibitem{he2022masked}
Kaiming He, Xinlei Chen, Saining Xie, Yanghao Li, Piotr Doll{\'a}r, and Ross Girshick.
\newblock Masked autoencoders are scalable vision learners.
\newblock In {\em Proceedings of the IEEE/CVF Conference on Computer Vision and Pattern Recognition}, pages 16000--16009, 2022.

\bibitem{he2016deep}
Kaiming He, Xiangyu Zhang, Shaoqing Ren, and Jian Sun.
\newblock Deep residual learning for image recognition.
\newblock In {\em Proceedings of the IEEE conference on computer vision and pattern recognition}, pages 770--778, 2016.

\bibitem{henriques2014high}
Jo{\~a}o~F Henriques, Rui Caseiro, Pedro Martins, and Jorge Batista.
\newblock High-speed tracking with kernelized correlation filters.
\newblock {\em IEEE transactions on pattern analysis and machine intelligence}, 37(3):583--596, 2014.

\bibitem{hu2021learning}
Li Hu, Peng Zhang, Bang Zhang, Pan Pan, Yinghui Xu, and Rong Jin.
\newblock Learning position and target consistency for memory-based video object segmentation.
\newblock In {\em Proceedings of the IEEE/CVF Conference on Computer Vision and Pattern Recognition}, pages 4144--4154, 2021.

\bibitem{huang2019got}
Lianghua Huang, Xin Zhao, and Kaiqi Huang.
\newblock Got-10k: A large high-diversity benchmark for generic object tracking in the wild.
\newblock {\em IEEE transactions on pattern analysis and machine intelligence}, 43(5):1562--1577, 2019.

\bibitem{kristan2022tenth}
Matej Kristan, Ale{\v{s}} Leonardis, Ji{\v{r}}{\'\i} Matas, Michael Felsberg, Roman Pflugfelder, Joni-Kristian K{\"a}m{\"a}r{\"a}inen, Hyung~Jin Chang, Martin Danelljan, Luka~{\v{C}}ehovin Zajc, Alan Luke{\v{z}}i{\v{c}}, et~al.
\newblock The tenth visual object tracking vot2022 challenge results.
\newblock In {\em European Conference on Computer Vision}, pages 431--460. Springer, 2022.

\bibitem{law2018cornernet}
Hei Law and Jia Deng.
\newblock Cornernet: Detecting objects as paired keypoints.
\newblock In {\em Proceedings of the European conference on computer vision (ECCV)}, pages 734--750, 2018.

\bibitem{li2019siamrpn++}
Bo Li, Wei Wu, Qiang Wang, Fangyi Zhang, Junliang Xing, and Junjie Yan.
\newblock Siamrpn++: Evolution of siamese visual tracking with very deep networks.
\newblock In {\em Proceedings of the IEEE/CVF conference on computer vision and pattern recognition}, pages 4282--4291, 2019.

\bibitem{li2018high}
Bo Li, Junjie Yan, Wei Wu, Zheng Zhu, and Xiaolin Hu.
\newblock High performance visual tracking with siamese region proposal network.
\newblock In {\em Proceedings of the IEEE conference on computer vision and pattern recognition}, pages 8971--8980, 2018.

\bibitem{li2023transformer}
Xiangtai Li, Henghui Ding, Wenwei Zhang, Haobo Yuan, Jiangmiao Pang, Guangliang Cheng, Kai Chen, Ziwei Liu, and Chen~Change Loy.
\newblock Transformer-based visual segmentation: A survey.
\newblock {\em arXiv preprint arXiv:2304.09854}, 2023.

\bibitem{li2023tube}
Xiangtai Li, Haobo Yuan, Wenwei Zhang, Guangliang Cheng, Jiangmiao Pang, and Chen~Change Loy.
\newblock Tube-link: A flexible cross tube baseline for universal video segmentation.
\newblock In {\em Proceedings of the IEEE/CVF International Conference on Computer Vision}, 2023.

\bibitem{li2022video}
Xiangtai Li, Wenwei Zhang, Jiangmiao Pang, Kai Chen, Guangliang Cheng, Yunhai Tong, and Chen~Change Loy.
\newblock Video k-net: A simple, strong, and unified baseline for video segmentation.
\newblock In {\em Proceedings of the IEEE/CVF Conference on Computer Vision and Pattern Recognition}, pages 18847--18857, 2022.

\bibitem{li2020fast}
Yu Li, Zhuoran Shen, and Ying Shan.
\newblock Fast video object segmentation using the global context module.
\newblock In {\em Computer Vision--ECCV 2020: 16th European Conference, Glasgow, UK, August 23--28, 2020, Proceedings, Part X 16}, pages 735--750. Springer, 2020.

\bibitem{liang2023local}
Chen Liang, Wenguan Wang, Tianfei Zhou, Jiaxu Miao, Yawei Luo, and Yi Yang.
\newblock Local-global context aware transformer for language-guided video segmentation.
\newblock {\em IEEE Transactions on Pattern Analysis and Machine Intelligence}, 45(8):10055--10069, 2023.

\bibitem{liang2020video}
Yongqing Liang, Xin Li, Navid Jafari, and Jim Chen.
\newblock Video object segmentation with adaptive feature bank and uncertain-region refinement.
\newblock {\em Advances in Neural Information Processing Systems}, 33:3430--3441, 2020.

\bibitem{lin2021swintrack}
Liting Lin, Heng Fan, Yong Xu, and Haibin Ling.
\newblock Swintrack: A simple and strong baseline for transformer tracking.
\newblock {\em arXiv preprint arXiv:2112.00995}, 2021.

\bibitem{lin2017feature}
Tsung-Yi Lin, Piotr Doll{\'a}r, Ross Girshick, Kaiming He, Bharath Hariharan, and Serge Belongie.
\newblock Feature pyramid networks for object detection.
\newblock In {\em Proceedings of the IEEE conference on computer vision and pattern recognition}, pages 2117--2125, 2017.

\bibitem{lin2014microsoft}
Tsung-Yi Lin, Michael Maire, Serge Belongie, James Hays, Pietro Perona, Deva Ramanan, Piotr Doll{\'a}r, and C~Lawrence Zitnick.
\newblock Microsoft coco: Common objects in context.
\newblock In {\em European conference on computer vision}, pages 740--755. Springer, 2014.

\bibitem{lu2020video}
Xiankai Lu, Wenguan Wang, Martin Danelljan, Tianfei Zhou, Jianbing Shen, and Luc Van~Gool.
\newblock Video object segmentation with episodic graph memory networks.
\newblock In {\em Computer Vision--ECCV 2020: 16th European Conference, Glasgow, UK, August 23--28, 2020, Proceedings, Part III 16}, pages 661--679. Springer, 2020.

\bibitem{luiten2019premvos}
Jonathon Luiten, Paul Voigtlaender, and Bastian Leibe.
\newblock Premvos: Proposal-generation, refinement and merging for video object segmentation.
\newblock In {\em Computer Vision--ACCV 2018: 14th Asian Conference on Computer Vision, Perth, Australia, December 2--6, 2018, Revised Selected Papers, Part IV}, pages 565--580. Springer, 2019.

\bibitem{lukezic2020d3s}
Alan Lukezic, Jiri Matas, and Matej Kristan.
\newblock D3s-a discriminative single shot segmentation tracker.
\newblock In {\em Proceedings of the IEEE/CVF conference on computer vision and pattern recognition}, pages 7133--7142, 2020.

\bibitem{lukezic2017discriminative}
Alan Lukezic, Tomas Vojir, Luka ˇCehovin~Zajc, Jiri Matas, and Matej Kristan.
\newblock Discriminative correlation filter with channel and spatial reliability.
\newblock In {\em Proceedings of the IEEE conference on computer vision and pattern recognition}, pages 6309--6318, 2017.

\bibitem{ma2022unified}
Fan Ma, Mike~Zheng Shou, Linchao Zhu, Haoqi Fan, Yilei Xu, Yi Yang, and Zhicheng Yan.
\newblock Unified transformer tracker for object tracking.
\newblock In {\em Proceedings of the IEEE/CVF Conference on Computer Vision and Pattern Recognition}, pages 8781--8790, 2022.

\bibitem{maninis2018video}
K-K Maninis, Sergi Caelles, Yuhua Chen, Jordi Pont-Tuset, Laura Leal-Taix{\'e}, Daniel Cremers, and Luc Van~Gool.
\newblock Video object segmentation without temporal information.
\newblock {\em IEEE transactions on pattern analysis and machine intelligence}, 41(6):1515--1530, 2018.

\bibitem{mao2021joint}
Yunyao Mao, Ning Wang, Wengang Zhou, and Houqiang Li.
\newblock Joint inductive and transductive learning for video object segmentation.
\newblock In {\em Proceedings of the IEEE/CVF international conference on computer vision}, pages 9670--9679, 2021.

\bibitem{mayer2022transforming}
Christoph Mayer, Martin Danelljan, Goutam Bhat, Matthieu Paul, Danda~Pani Paudel, Fisher Yu, and Luc Van~Gool.
\newblock Transforming model prediction for tracking.
\newblock In {\em Proceedings of the IEEE/CVF conference on computer vision and pattern recognition}, pages 8731--8740, 2022.

\bibitem{mayer2021learning}
Christoph Mayer, Martin Danelljan, Danda~Pani Paudel, and Luc Van~Gool.
\newblock Learning target candidate association to keep track of what not to track.
\newblock In {\em Proceedings of the IEEE/CVF International Conference on Computer Vision}, pages 13444--13454, 2021.

\bibitem{meinhardt2020make}
Tim Meinhardt and Laura Leal-Taix{\'e}.
\newblock Make one-shot video object segmentation efficient again.
\newblock {\em Advances in Neural Information Processing Systems}, 33:10607--10619, 2020.

\bibitem{milan2016mot16}
Anton Milan, Laura Leal-Taix{\'e}, Ian Reid, Stefan Roth, and Konrad Schindler.
\newblock Mot16: A benchmark for multi-object tracking.
\newblock {\em arXiv preprint arXiv:1603.00831}, 2016.

\bibitem{muller2018trackingnet}
Matthias Muller, Adel Bibi, Silvio Giancola, Salman Alsubaihi, and Bernard Ghanem.
\newblock Trackingnet: A large-scale dataset and benchmark for object tracking in the wild.
\newblock In {\em Proceedings of the European conference on computer vision (ECCV)}, pages 300--317, 2018.

\bibitem{nam2016learning}
Hyeonseob Nam and Bohyung Han.
\newblock Learning multi-domain convolutional neural networks for visual tracking.
\newblock In {\em Proceedings of the IEEE conference on computer vision and pattern recognition}, pages 4293--4302, 2016.

\bibitem{oh2018fast}
Seoung~Wug Oh, Joon-Young Lee, Kalyan Sunkavalli, and Seon~Joo Kim.
\newblock Fast video object segmentation by reference-guided mask propagation.
\newblock In {\em Proceedings of the IEEE conference on computer vision and pattern recognition}, pages 7376--7385, 2018.

\bibitem{oh2019video}
Seoung~Wug Oh, Joon-Young Lee, Ning Xu, and Seon~Joo Kim.
\newblock Video object segmentation using space-time memory networks.
\newblock In {\em Proceedings of the IEEE/CVF International Conference on Computer Vision}, pages 9226--9235, 2019.

\bibitem{papadopoulos2017extreme}
Dim~P Papadopoulos, Jasper~RR Uijlings, Frank Keller, and Vittorio Ferrari.
\newblock Extreme clicking for efficient object annotation.
\newblock In {\em Proceedings of the IEEE international conference on computer vision}, pages 4930--4939, 2017.

\bibitem{paul2022robust}
Matthieu Paul, Martin Danelljan, Christoph Mayer, and Luc Van~Gool.
\newblock Robust visual tracking by segmentation.
\newblock In {\em Computer Vision--ECCV 2022: 17th European Conference, Tel Aviv, Israel, October 23--27, 2022, Proceedings, Part XXII}, pages 571--588. Springer, 2022.

\bibitem{pont20172017}
Jordi Pont-Tuset, Federico Perazzi, Sergi Caelles, Pablo Arbel{\'a}ez, Alex Sorkine-Hornung, and Luc Van~Gool.
\newblock The 2017 davis challenge on video object segmentation.
\newblock {\em arXiv preprint arXiv:1704.00675}, 2017.

\bibitem{rezatofighi2019generalized}
Hamid Rezatofighi, Nathan Tsoi, JunYoung Gwak, Amir Sadeghian, Ian Reid, and Silvio Savarese.
\newblock Generalized intersection over union: A metric and a loss for bounding box regression.
\newblock In {\em Proceedings of the IEEE/CVF conference on computer vision and pattern recognition}, pages 658--666, 2019.

\bibitem{sandler2018mobilenetv2}
Mark Sandler, Andrew Howard, Menglong Zhu, Andrey Zhmoginov, and Liang-Chieh Chen.
\newblock Mobilenetv2: Inverted residuals and linear bottlenecks.
\newblock In {\em Proceedings of the IEEE conference on computer vision and pattern recognition}, pages 4510--4520, 2018.

\bibitem{seong2021hierarchical}
Hongje Seong, Seoung~Wug Oh, Joon-Young Lee, Seongwon Lee, Suhyeon Lee, and Euntai Kim.
\newblock Hierarchical memory matching network for video object segmentation.
\newblock In {\em Proceedings of the IEEE/CVF International Conference on Computer Vision}, pages 12889--12898, 2021.

\bibitem{shi2015hierarchical}
Jianping Shi, Qiong Yan, Li Xu, and Jiaya Jia.
\newblock Hierarchical image saliency detection on extended cssd.
\newblock {\em IEEE transactions on pattern analysis and machine intelligence}, 38(4):717--729, 2015.

\bibitem{shin2017pixel}
Jae Shin~Yoon, Francois Rameau, Junsik Kim, Seokju Lee, Seunghak Shin, and In So~Kweon.
\newblock Pixel-level matching for video object segmentation using convolutional neural networks.
\newblock In {\em Proceedings of the IEEE international conference on computer vision}, pages 2167--2176, 2017.

\bibitem{song2022transformer}
Zikai Song, Junqing Yu, Yi-Ping~Phoebe Chen, and Wei Yang.
\newblock Transformer tracking with cyclic shifting window attention.
\newblock In {\em Proceedings of the IEEE/CVF conference on computer vision and pattern recognition}, pages 8791--8800, 2022.

\bibitem{valmadre2017end}
Jack Valmadre, Luca Bertinetto, Joao Henriques, Andrea Vedaldi, and Philip~HS Torr.
\newblock End-to-end representation learning for correlation filter based tracking.
\newblock In {\em Proceedings of the IEEE conference on computer vision and pattern recognition}, pages 2805--2813, 2017.

\bibitem{vaswani2017attention}
Ashish Vaswani, Noam Shazeer, Niki Parmar, Jakob Uszkoreit, Llion Jones, Aidan~N Gomez, {\L}ukasz Kaiser, and Illia Polosukhin.
\newblock Attention is all you need.
\newblock {\em Advances in neural information processing systems}, 30, 2017.

\bibitem{voigtlaender2019feelvos}
Paul Voigtlaender, Yuning Chai, Florian Schroff, Hartwig Adam, Bastian Leibe, and Liang-Chieh Chen.
\newblock Feelvos: Fast end-to-end embedding learning for video object segmentation.
\newblock In {\em Proceedings of the IEEE/CVF Conference on Computer Vision and Pattern Recognition}, pages 9481--9490, 2019.

\bibitem{voigtlaender2019mots}
Paul Voigtlaender, Michael Krause, Aljosa Osep, Jonathon Luiten, Berin Balachandar~Gnana Sekar, Andreas Geiger, and Bastian Leibe.
\newblock Mots: Multi-object tracking and segmentation.
\newblock In {\em Proceedings of the ieee/cvf conference on computer vision and pattern recognition}, pages 7942--7951, 2019.

\bibitem{voigtlaender2017online}
Paul Voigtlaender and Bastian Leibe.
\newblock Online adaptation of convolutional neural networks for video object segmentation.
\newblock {\em arXiv preprint arXiv:1706.09364}, 2017.

\bibitem{voigtlaender2020siam}
Paul Voigtlaender, Jonathon Luiten, Philip~HS Torr, and Bastian Leibe.
\newblock Siam r-cnn: Visual tracking by re-detection.
\newblock In {\em Proceedings of the IEEE/CVF conference on computer vision and pattern recognition}, pages 6578--6588, 2020.

\bibitem{wang2020tracking}
Guangting Wang, Chong Luo, Xiaoyan Sun, Zhiwei Xiong, and Wenjun Zeng.
\newblock Tracking by instance detection: A meta-learning approach.
\newblock In {\em Proceedings of the IEEE/CVF conference on computer vision and pattern recognition}, pages 6288--6297, 2020.

\bibitem{wang2021transformer}
Ning Wang, Wengang Zhou, Jie Wang, and Houqiang Li.
\newblock Transformer meets tracker: Exploiting temporal context for robust visual tracking.
\newblock In {\em Proceedings of the IEEE/CVF Conference on Computer Vision and Pattern Recognition}, pages 1571--1580, 2021.

\bibitem{wang2019fast}
Qiang Wang, Li Zhang, Luca Bertinetto, Weiming Hu, and Philip~HS Torr.
\newblock Fast online object tracking and segmentation: A unifying approach.
\newblock In {\em Proceedings of the IEEE/CVF conference on Computer Vision and Pattern Recognition}, pages 1328--1338, 2019.

\bibitem{wang2021survey}
Wenguan Wang, Tianfei Zhou, Fatih Porikli, David Crandall, and Luc Van~Gool.
\newblock A survey on deep learning technique for video segmentation.
\newblock {\em arXiv e-prints}, pages arXiv--2107, 2021.

\bibitem{wang2019ranet}
Ziqin Wang, Jun Xu, Li Liu, Fan Zhu, and Ling Shao.
\newblock Ranet: Ranking attention network for fast video object segmentation.
\newblock In {\em Proceedings of the IEEE/CVF International Conference on Computer Vision}, pages 3978--3987, 2019.

\bibitem{wang2021different}
Zhongdao Wang, Hengshuang Zhao, Ya-Li Li, Shengjin Wang, Philip Torr, and Luca Bertinetto.
\newblock Do different tracking tasks require different appearance models?
\newblock {\em Advances in Neural Information Processing Systems}, 34:726--738, 2021.

\bibitem{wu2021cvt}
Haiping Wu, Bin Xiao, Noel Codella, Mengchen Liu, Xiyang Dai, Lu Yuan, and Lei Zhang.
\newblock Cvt: Introducing convolutions to vision transformers.
\newblock In {\em Proceedings of the IEEE/CVF International Conference on Computer Vision}, pages 22--31, 2021.

\bibitem{xie2022correlation}
Fei Xie, Chunyu Wang, Guangting Wang, Yue Cao, Wankou Yang, and Wenjun Zeng.
\newblock Correlation-aware deep tracking.
\newblock In {\em Proceedings of the IEEE/CVF Conference on Computer Vision and Pattern Recognition}, pages 8751--8760, 2022.

\bibitem{xu2018youtube}
Ning Xu, Linjie Yang, Yuchen Fan, Dingcheng Yue, Yuchen Liang, Jianchao Yang, and Thomas Huang.
\newblock Youtube-vos: A large-scale video object segmentation benchmark.
\newblock {\em arXiv preprint arXiv:1809.03327}, 2018.

\bibitem{xu2023video}
Yuanyou Xu, Zongxin Yang, and Yi Yang.
\newblock Video object segmentation in panoptic wild scenes.
\newblock In {\em International Joint Conference on Artificial Intelligence}, 2023.

\bibitem{yan2022towards}
Bin Yan, Yi Jiang, Peize Sun, Dong Wang, Zehuan Yuan, Ping Luo, and Huchuan Lu.
\newblock Towards grand unification of object tracking.
\newblock In {\em Computer Vision--ECCV 2022: 17th European Conference, Tel Aviv, Israel, October 23--27, 2022, Proceedings, Part XXI}, pages 733--751. Springer, 2022.

\bibitem{yan2023universal}
Bin Yan, Yi Jiang, Jiannan Wu, Dong Wang, Ping Luo, Zehuan Yuan, and Huchuan Lu.
\newblock Universal instance perception as object discovery and retrieval.
\newblock In {\em Proceedings of the IEEE/CVF Conference on Computer Vision and Pattern Recognition}, pages 15325--15336, 2023.

\bibitem{yan2021learning}
Bin Yan, Houwen Peng, Jianlong Fu, Dong Wang, and Huchuan Lu.
\newblock Learning spatio-temporal transformer for visual tracking.
\newblock In {\em Proceedings of the IEEE/CVF international conference on computer vision}, pages 10448--10457, 2021.

\bibitem{yan2021alpha}
Bin Yan, Xinyu Zhang, Dong Wang, Huchuan Lu, and Xiaoyun Yang.
\newblock Alpha-refine: Boosting tracking performance by precise bounding box estimation.
\newblock In {\em Proceedings of the IEEE/CVF Conference on Computer Vision and Pattern Recognition}, pages 5289--5298, 2021.

\bibitem{yang2018efficient}
Linjie Yang, Yanran Wang, Xuehan Xiong, Jianchao Yang, and Aggelos~K Katsaggelos.
\newblock Efficient video object segmentation via network modulation.
\newblock In {\em Proceedings of the IEEE Conference on Computer Vision and Pattern Recognition}, pages 6499--6507, 2018.

\bibitem{mkr}
Yi Yang, Yueting Zhuang, and Yunhe Pan.
\newblock Multiple knowledge representation for big data artificial intelligence: framework, applications, and case studies.
\newblock {\em Frontiers of Information Technology \& Electronic Engineering}, 22(12):1551--1558, 2021.

\bibitem{yang2022associating}
Zongxin Yang, Xiaohan Wang, Jiaxu Miao, Yunchao Wei, Wenguan Wang, and Yi Yang.
\newblock Scalable video object segmentation with identification mechanism.
\newblock {\em arXiv preprint arXiv:2203.11442}, 2023.

\bibitem{yang2020collaborative}
Zongxin Yang, Yunchao Wei, and Yi Yang.
\newblock Collaborative video object segmentation by foreground-background integration.
\newblock In {\em European Conference on Computer Vision}, pages 332--348. Springer, 2020.

\bibitem{yang2021associating}
Zongxin Yang, Yunchao Wei, and Yi Yang.
\newblock Associating objects with transformers for video object segmentation.
\newblock {\em Advances in Neural Information Processing Systems}, 34, 2021.

\bibitem{yang2021collaborative}
Zongxin Yang, Yunchao Wei, and Yi Yang.
\newblock Collaborative video object segmentation by multi-scale foreground-background integration.
\newblock {\em IEEE Transactions on Pattern Analysis and Machine Intelligence}, 2021.

\bibitem{yang2022decoupling}
Zongxin Yang and Yi Yang.
\newblock Decoupling features in hierarchical propagation for video object segmentation.
\newblock {\em Advances in Neural Information Processing Systems}, 2022.

\bibitem{yang2021towards}
Zongxin Yang, Jian Zhang, Wenhao Wang, Wenhua Han, Yue Yu, Yingying Li, Jian Wang, Yunchao Wei, Yifan Sun, and Yi Yang.
\newblock Towards multi-object association from foreground-background integration.
\newblock In {\em CVPR Workshops}, volume~2, 2021.

\bibitem{ye2022joint}
Botao Ye, Hong Chang, Bingpeng Ma, Shiguang Shan, and Xilin Chen.
\newblock Joint feature learning and relation modeling for tracking: A one-stream framework.
\newblock In {\em Computer Vision--ECCV 2022: 17th European Conference, Tel Aviv, Israel, October 23--27, 2022, Proceedings, Part XXII}, pages 341--357. Springer, 2022.

\bibitem{yu2020deformable}
Yuechen Yu, Yilei Xiong, Weilin Huang, and Matthew~R Scott.
\newblock Deformable siamese attention networks for visual object tracking.
\newblock In {\em Proceedings of the IEEE/CVF conference on computer vision and pattern recognition}, pages 6728--6737, 2020.

\bibitem{zhang2023boosting}
Yurong Zhang, Liulei Li, Wenguan Wang, Rong Xie, Li Song, and Wenjun Zhang.
\newblock Boosting video object segmentation via space-time correspondence learning.
\newblock In {\em Proceedings of the IEEE/CVF Conference on Computer Vision and Pattern Recognition}, pages 2246--2256, 2023.

\bibitem{zhang2020ocean}
Zhipeng Zhang, Houwen Peng, Jianlong Fu, Bing Li, and Weiming Hu.
\newblock Ocean: Object-aware anchor-free tracking.
\newblock In {\em Computer Vision--ECCV 2020: 16th European Conference, Glasgow, UK, August 23--28, 2020, Proceedings, Part XXI 16}, pages 771--787. Springer, 2020.

\bibitem{zhao2021generating}
Bin Zhao, Goutam Bhat, Martin Danelljan, Luc Van~Gool, and Radu Timofte.
\newblock Generating masks from boxes by mining spatio-temporal consistencies in videos.
\newblock In {\em Proceedings of the IEEE/CVF International Conference on Computer Vision}, pages 13556--13566, 2021.

\bibitem{zhou2019bottom}
Xingyi Zhou, Jiacheng Zhuo, and Philipp Krahenbuhl.
\newblock Bottom-up object detection by grouping extreme and center points.
\newblock In {\em Proceedings of the IEEE/CVF conference on computer vision and pattern recognition}, pages 850--859, 2019.

\end{thebibliography}
}
\end{document}